\documentclass{ieeeaccess}
\usepackage{cite}
\usepackage{amsmath,amssymb,amsfonts}
\usepackage{algorithmic}
\usepackage{graphicx}
\usepackage{textcomp}
\usepackage{booktabs}
\usepackage{array}
\usepackage{tabularx}
\usepackage{hyperref}
\hypersetup{hidelinks}
\usepackage{xcolor}
\expandafter\def\expandafter\UrlBreaks\expandafter{\UrlBreaks\do\-}

\def\BibTeX{{\rm B\kern-.05em{\sc i\kern-.025em b}\kern-.08em
    T\kern-.1667em\lower.7ex\hbox{E}\kern-.125emX}}

\begin{document}

\history{Author preprint, November 2026 (not peer reviewed).}
\doi{}

\title{An Interaction-Aware Decision Framework for FDIC-Assisted
Distressed Bank Acquisitions: Empirical Evidence from the
FDIC Failed-Bank Panel and the SVB--First Citizens Case}
\author{\uppercase{Yousef Nejatbakhsh}\authorrefmark{1}, \uppercase{Malihe Aliasgari}\authorrefmark{2}, \IEEEmembership{Member, IEEE}, \uppercase{Parsa Safaee}\authorrefmark{3}, \uppercase{Sina Shokoohyar}\authorrefmark{4}}
\address[1]{Department of Statistics, Rutgers University, New Brunswick, NJ 08901, USA (e-mail: nejatbakhsh.y@gmail.com)}
\address[2]{Department of Mathematical Sciences, Kean University, Union, NJ 07083, USA (e-mail: malihe.aliasgari@kean.edu)}
\address[3]{Eindhoven Artificial Intelligence Systems Institute (EAISI), Eindhoven University of Technology, Eindhoven, The Netherlands (e-mail: p.safaee@tue.nl)}
\address[4]{Department of Computing and Decision Sciences
Stillman School of Business
Seton Hall University
Jubilee Hall 615
400 South Orange Avenue
South Orange, NJ. 07079
(e-mail: sina.shokoohyar@shu.edu)}

\tfootnote{This work received no external funding. The authors declare
no conflicts of interest.}

\markboth
{Nejatbakhsh \textit{et al.}: Spider Network Filtering for Distressed Bank Acquisitions}
{Nejatbakhsh \textit{et al.}: Spider Network Filtering for Distressed Bank Acquisitions}

\corresp{Corresponding author: Malihe Aliasgari
(e-mail: malihe.aliasgari@kean.edu).}

\begin{abstract}
Acquirers of failed banks weigh acquisition value, regulatory
protection, risk, buyer strength, and execution difficulty jointly,
because each signal's meaning depends on the others: high
non-performing loans matter less when government loss sharing
absorbs the losses. The proposed \emph{Spider Network Filtering
Framework} encodes such dependencies as four interpretable interaction
rules among five dimensions, yielding graph-inspired features for a
logistic model and a Pursue/Caution/Avoid decision rule. We evaluate
it on 55 acquisitions assisted by the Federal Deposit Insurance
Corporation in 2008--2024 by acquirers with a parent listed in the
United States. Six financial inputs come from regulatory reports
published before each failure; two deal terms, which the winning
bidder knows when it bids but investors learn at the announcement,
come from the regulator's press releases. The outcome, a positive
three-day abnormal return around the announcement, records the
market's initial reception, not realized success. Under grouped,
nested cross-validation, the Spider-enhanced model reaches a mean area
under the receiver operating characteristic curve of 0.594,
against 0.615 for logistic regression on the same inputs and
0.632 for ridge regression; permutation tests that repeat the
full model selection give at most weak evidence that any of them ranks
better than chance. The Spider score used directly does not rank
better than chance, and an apparent link between target distress and
higher returns disappears against a bank-sector benchmark. Thresholds
tuned for the F1 score approve most acquisitions with negative returns
and realize negative utility when each costs three times what a
positively received acquisition gains.
\end{abstract}

\begin{keywords}
Bank failure, distressed mergers and acquisitions, event study,
explainable AI, FDIC, interaction rules, logistic regression,
reproducibility, risk analytics, Silicon Valley Bank
\end{keywords}

\titlepgskip=-15pt
\maketitle

\section{Introduction}
\label{sec:intro}

Distressed bank acquisitions occupy a unique position in financial
decision-making. Unlike standard mergers and acquisitions (M\&A), they
combine the urgency of regulatory intervention, the opacity of
distressed balance sheets, and the complexity of government
loss-sharing arrangements into a single time-constrained decision.
When the Federal Deposit Insurance Corporation (FDIC) markets a
failing bank, prospective acquirers typically have a short window of
confidential due diligence, often a few weeks, to decide whether to
bid and on what terms. The acquisition of
Silicon Valley Bank (SVB) by First Citizens BancShares in March 2023
exemplifies this dynamic. SVB's collapse, at the time the
second-largest bank failure in U.S. history, was precipitated by a classic liquidity
crisis: concentrated deposit outflows among interconnected
venture-capital-backed firms, compounded by unrealized losses in a
long-duration securities portfolio accumulated during the low-rate era
of 2020--2021. The FDIC subsequently facilitated the acquisition,
transferring approximately
\$72 billion in assets, purchased at a discount of about 23\%, and
\$56 billion in deposits to First Citizens~\cite{fdic2023svb,
firstcitizens2023}.

The academic literature on bank distress prediction is extensive,
drawing on Altman's Z-score~\cite{altman1968}, logistic
regression~\cite{martin1977}, and more recently machine learning
approaches including random forests, gradient boosting, and neural
networks~\cite{jones2017}. However, the literature on
\emph{acquisition decision-making} in the context of distressed banks
is considerably thinner. Existing frameworks tend to evaluate
acquisition targets through isolated financial ratios rather than
through the connected logic that practitioners apply. A high
non-performing loan ratio, for example, is not independently
informative: it matters less when FDIC loss sharing absorbs the credit
losses, and more when the acquirer's capital is thin. Standard models
do not capture these interdependencies explicitly.

This paper makes three contributions. First, we shift the modeling
objective from bank-failure \emph{prediction} to post-failure
acquisition \emph{decision support}, a distinct and comparatively
under-studied problem. Second, we formalize the Spider Network
Filtering Framework (SNFF), an interpretable, graph-inspired
feature-engineering scheme that encodes the conditional dependencies
among credit risk, regulatory protection, liquidity distress,
structural deterioration, buyer capital, and execution burden as four
explicit interaction rules, and we build a panel of 55 FDIC-assisted
acquisitions by U.S.-listed acquirers in which the six financial
inputs come from call reports published before the failure, the two
deal terms are coded from the FDIC press release, the information
available to the bidder when it bids is stated explicitly, and the
event day of every outcome is fixed from the verified timing of the
announcement. Third, we evaluate the framework against seven
alternatives under grouped, nested cross-validation, with uncertainty
assessed by procedures that repeat the model fitting and selection,
and report what the evidence supports: the Spider features do not
improve ranking relative to logistic models on the same inputs and
change calibration only slightly, in a direction that depends on the
outcome window; the evidence that any model discriminates better than
chance is weak at this sample size; and an apparent association
between the target's distress and the acquirer's announcement return,
which would reverse the score's risk penalty, holds against the
S\&P~500 but not against a bank-sector benchmark. A
prescriptive analysis shows that decision thresholds must reflect the
asymmetric cost of acquisition errors rather than classification
accuracy, and that under such costs no policy evaluated here improves
materially on declining every transaction.

The central research question of this study is: \textit{Do
interaction-aware Spider Network features improve the ranking and
calibration of distressed bank acquisitions by the sign of the
acquirer's announcement return, relative to traditional
financial-ratio models, and can four domain-motivated
interaction rules match unstructured interaction and machine-learning
models?} The ablation study and baseline comparisons in
Section~\ref{sec:results} answer both parts. We state the first part
as a testable hypothesis:

\begin{quote}
\textbf{H1}: The Spider Network Filter generates interaction-based
features that improve ranking performance, measured by the area under
the receiver operating characteristic curve (AUC-ROC, or AUC), and probability
calibration (Brier score) relative to standard logistic regression
using the measured inputs only (Model~2).
\end{quote}

\noindent The evidence in Section~\ref{sec:results} does not support
H1.

The remainder of the paper is organized as follows.
Section~\ref{sec:literature} reviews the relevant literature on bank
distress, M\&A decision frameworks, and network-based risk models.
Section~\ref{sec:framework} presents the Spider Network Filtering
Framework, including the component score formulas and interaction
rules. Section~\ref{sec:models} specifies all four models and their
evaluation criteria. Section~\ref{sec:data} describes the data
structure and variable definitions. Section~\ref{sec:svb} applies the
framework to the SVB--First Citizens case. Section~\ref{sec:results}
reports the model comparisons, robustness checks, and decision
analysis. Section~\ref{sec:discussion} discusses the findings,
including why the data cannot settle the direction of the risk
penalty, and their implications for practitioners and regulators. Section~\ref{sec:conclusion}
concludes with recommendations for future research.

\section{Literature Review}
\label{sec:literature}

\subsection{Bank Distress Prediction}

The prediction of bank failure has been a central problem in financial
economics since the banking crises of the 1970s and 1980s. Early
approaches relied on univariate financial ratio analysis and the CAMELS
framework (Capital adequacy, Asset quality, Management, Earnings,
Liquidity, Sensitivity), which remains the supervisory standard for
U.S. bank examination~\cite{cole1995}. Logistic regression models
became the dominant statistical tool following Martin~\cite{martin1977}
and Pettway and Sinkey~\cite{pettway1980}, who demonstrated the
predictive power of balance sheet ratios in identifying failing
institutions. More recent work has expanded the predictor set to
include macroeconomic variables, network centrality measures, and
text-derived sentiment signals~\cite{kou2021}.

\subsection{Distressed Mergers and Acquisitions}

The M\&A literature on distressed targets is distinct from the broader
acquisition literature in several important respects. Distressed
acquisitions typically occur at substantial discounts to book value,
involve regulatory intermediaries, and create asymmetric information
problems that amplify integration risk~\cite{meierservaes2014}. The FDIC
assisted-transaction framework, which may include loss-sharing
agreements, purchase-and-assumption structures, and deposit
transfer arrangements, introduces a regulatory protection dimension
that has no equivalent in standard M\&A analysis~\cite{james1991}.
Prior studies on failed-bank resolution and acquirer outcomes provide
mixed evidence on whether FDIC-assisted acquisitions create sustained
value for acquirers: the broader literature on distressed-asset
acquisitions documents excess returns to acquirers of distressed and
bankrupt firms~\cite{meierservaes2014}, while studies of bank scale
economies suggest integration costs can erode these gains over time
for some acquirers~\cite{wheelock2012}.

A more direct literature studies the acquirers of failed banks. James
and Wier~\cite{james1987} find that winning bidders in FDIC
failed-bank auctions earn significant positive abnormal returns,
consistent with wealth transfers from the FDIC to acquirers, and
Giliberto and Varaiya~\cite{giliberto1989} show, for 322 FDIC auctions
between 1965 and 1985, that winning bids rise with the number of
competing bidders, consistent with bidders failing to adjust for the
winner's curse. For the 2008--2013 failure wave,
Cowan and Salotti~\cite{cowan2015} report substantial positive
abnormal returns to winning bidders that decline with the amount bid,
and identify FDIC loss-share guarantees as an important determinant of
bids. Granja, Matvos, and Seru~\cite{granja2017} show that failed
banks are sold mostly to nearby acquirers with similar loan portfolios
and that weak capitalization limits which banks can bid, which links
the pool of bidders to the losses the FDIC realizes. The predominance
of positive announcement returns in our panel (35 of 55 transactions)
is consistent with this evidence. These studies explain average
acquirer returns and auction outcomes; we ask instead whether the
value of an individual transaction can be ranked in advance from the
signals available when a bid is prepared.

\subsection{Multi-Criteria Decision Models in Finance}

Weighted scoring models and multi-criteria decision analysis (MCDA)
have a long history in financial decision-making, particularly in
credit rating, supplier selection, and investment
screening~\cite{hwang1981, saaty1980}. These models offer high
interpretability but suffer from two well-known weaknesses: subjective
weight assignment and the inability to capture interaction effects
among criteria. Hybrid models that combine MCDA with statistical or
machine learning methods have been proposed to address these
weaknesses~\cite{Kou2014}, but applications to distressed bank
acquisitions remain scarce.

\subsection{Network Models in Financial Risk}

Graph-theoretic and network-based approaches to financial risk have
grown substantially since the 2008 global financial crisis revealed the
importance of systemic interconnectedness~\cite{acemoglu2015,
allen2000}. Network models have been applied to interbank contagion,
deposit concentration risk, and counterparty exposure
mapping~\cite{braverman2020}. The application of network logic to
individual acquisition decision-making---rather than to systemic
risk---is a relatively unexplored direction. Our Spider Network
Filtering Framework borrows the language of interaction graphs from this
tradition to describe a transaction-level decision rule; unlike network
models of systemic risk, it does not compute on network structure.

\subsection{Deposit Network Contagion}

A separate and directly relevant literature documents how depositor
interconnectedness and coordination accelerate bank runs. Egan,
Horta\c{c}su, and Matvos~\cite{egan2017} show that uninsured
depositors above the FDIC insurance limit discipline weak banks by
withdrawing funds, a behavior directly relevant to concentrated
institutional deposit bases such as SVB's. Cookson, Fox, Gil-Bazo,
Imbet, and Schiller~\cite{cookson2023} provide direct empirical
evidence that social-media activity amplified the speed of the
deposit run that preceded SVB's and Signature Bank's failures in March
2023, demonstrating that depositor interconnectedness determines
contagion speed in a manner not captured by standard balance-sheet
ratios. Eisenberg and Noe~\cite{eisenberg2001} and Rogers and
Veraart~\cite{rogers2013} formalize how network topology determines
the scope of contagion in interbank systems more generally. The
Spider Network's deposit-outflow indicator and interaction rules are
a simplified, transaction-level analogue of this broader contagion
literature; we do not construct an explicit depositor graph in this
paper, and doing so---for example, via a Herfindahl-Hirschman index of
depositor concentration---is identified as a direction for future work
in Section~\ref{sec:discussion}.

\subsection{Explainable Machine Learning in Finance}

Recent work has emphasized the importance of explainability in
financial machine learning~\cite{Arrieta2020}. SHAP (SHapley Additive
exPlanations) values~\cite{lundberg2017} have been used to decompose
model predictions into variable-level contributions, enabling
regulators and practitioners to understand why a model recommends a
particular action. The integration of explainability methods with
distressed acquisition decision models is a natural extension that the
current paper begins to address through the Spider Network's
transparent interaction architecture.

Recent research has explored hybrid machine-learning
architectures for financial-market prediction. Nejatbakhsh and
Aliasgari~\cite{nejatbakhsh2026hybrid} combined long short-term memory
(LSTM) networks, transformer attention, sentiment analysis, and
federated learning for stock-market forecasting. In contrast,
the present study focuses on interpretable financial interactions
and the evaluation of decision-support models for FDIC-assisted
distressed bank acquisitions.

\section{The Spider Network Filtering Framework}
\label{sec:framework}

\subsection{Conceptual Foundation}

The Spider Network Filtering Framework is motivated by the observation
that distressed acquisition decisions are relational: the significance
of each financial signal depends on the values of connected signals.
A high non-performing loan (NPL) ratio is less threatening to an
acquirer when FDIC loss sharing covers the loans than when the
acquirer bears the credit losses alone. A deposit outflow before
failure points to a liquidity problem that a well-funded acquirer may
be able to reverse, unless the loan book is deteriorating at the same
time. A weakly capitalized buyer is more exposed than a strong one to
a target that is large relative to itself or whose credit quality is
declining. These dependencies constitute a network of conditional
relationships that simple additive scoring models cannot represent.

We summarize these dependencies in a conceptual interaction graph
$G = (V, E, W)$, where
\begin{align*}
V = \{&\text{Value},\ \text{Protection},\ \text{Risk},\\
      &\text{BuyerStrength},\ \text{Execution}\}
\end{align*}
is the set of five financial dimensions, $E \subseteq V \times V$ is
the set of directed edges, each standing for one interaction rule, and
$W: E \rightarrow \mathbb{R}$ assigns each rule its score adjustment.
The edges are
\begin{align}
E = \{&(\text{Risk},\ \text{Protection}),\ (\text{Risk},\ \text{Risk}),\notag\\
      &(\text{Risk},\ \text{BuyerStrength}),\notag\\
      &(\text{Execution},\ \text{BuyerStrength})\},
\end{align}
corresponding to the rules $Q$, $U$, $Z$, and $K$ of
Section~\ref{sec:interaction}. The edge from Risk to itself denotes an
interaction between two signals within the Risk dimension: rule~$U$
combines a deposit outflow with the absence of credit deterioration.
The graph documents which dimensions interact and by how much; no part
of the framework learns from the graph or computes on its structure,
and no propagation, centrality, or spectral quantity is used. The
framework is therefore best described as an interpretable,
graph-inspired feature-engineering scheme: weighted component scores
combined with four deterministic interaction rules.

\subsection{Component Score Formulas}
\label{sec:components}

Each acquisition candidate $i$ receives five component scores,
computed from two kinds of information (Section~\ref{sec:infoset}):
financial data published in the call reports of the failed and
acquiring banks before the announcement, and two terms of the
transaction---loss sharing and government support---that the bidder
knows when it bids and the market learns from the FDIC's
announcement. Deal-level
purchase discounts are not disclosed consistently across transactions,
so the Value dimension rests on the failed bank's deposit franchise,
and asset quality enters once, through the NPL ratio in the Risk
dimension. All continuous inputs are min-max normalized to $[0, 1]$,
with bounds estimated on the training data (Section~\ref{sec:leakage}).
The weights below are expert-assigned rather than estimated from data
or elicited through a formal procedure such as the Analytic Hierarchy
Process; none of the weights, adjustment sizes, or rule definitions was
estimated from the outcomes analyzed in this paper.

\subsubsection{Value Score}

The Value Score measures the franchise the acquirer obtains:
\begin{equation}
V_i = G_i,
\label{eq:value}
\end{equation}
where $G_i$ is the normalized deposit base of the failed bank in its
last public call report.

\subsubsection{Protection Score}

The Protection Score $P_i$ measures the extent to which regulatory
and governmental arrangements reduce downside risk:
\begin{equation}
P_i = 0.60\, F_i + 0.40\, H_i,
\label{eq:protection}
\end{equation}
where $F_i \in \{0, 1\}$ indicates an FDIC loss-share agreement in the
transaction and $H_i \in \{0, 1\}$ indicates extraordinary government
support (a bridge bank or a systemic-risk exception); both are terms
of the transaction (Section~\ref{sec:infoset}).

\subsubsection{Risk Score}

The Risk Score $R_i$ aggregates credit risk, liquidity distress, and
structural deterioration of the failed bank:
\begin{equation}
R_i = 0.40\, N_i + 0.25\, L_i + 0.35\, T_i,
\label{eq:risk}
\end{equation}
where $N_i$ is the normalized NPL ratio; $L_i \in \{0, 1\}$ equals 1
when the bank's deposit change over the two quarters ending at its last
public call report lies in the bottom quartile of the training data (a
deposit outflow); and $T_i \in \{0, 1\}$ equals 1 when the change in
its NPL ratio over the same two quarters lies in the top quartile (a
deteriorating loan book).

\subsubsection{Buyer Strength Score}

The Buyer Strength Score $B_i$ captures the acquiring institution's
capital adequacy:
\begin{equation}
B_i = C_i,
\label{eq:buyer}
\end{equation}
where $C_i$ is the normalized equity-to-assets ratio of the acquiring
bank at the last quarter-end at least 90 days before the failure.

\subsubsection{Execution Difficulty Score}

The Execution Difficulty Score $E_i$ proxies the integration burden by
the size of the target relative to the acquirer:
\begin{equation}
E_i = M_i,
\label{eq:exec}
\end{equation}
where $M_i$ is the normalized ratio of the failed bank's total assets
to the acquiring bank's total assets, winsorized at the 95th
percentile of the training data.

\subsection{Raw Spider Deal Score}
\label{sec:rawscore}

The raw Spider Deal Score $S_i$ combines the five component scores
into a single measure of acquisition attractiveness:
\begin{equation}
S_i = 0.30\, V_i + 0.20\, P_i + 0.20\, B_i - 0.20\, R_i - 0.10\, E_i
\label{eq:raw}
\end{equation}
The positive components (Value, Protection, Buyer Strength) increase
attractiveness; the negative components (Risk, Execution Difficulty)
reduce it. The weight constraint $\sum |w_k| = 1$ is satisfied.

\subsection{Spider Network Interaction Rules}
\label{sec:interaction}

The interaction rules distinguish the Spider Network from a simple
additive model. They encode the conditional dependencies among
dimensions identified in Section~\ref{sec:components}.

\begin{table}[!t]
\caption{Spider Network Interaction Rules}
\label{tab:interactions}
\centering
\footnotesize
\begin{tabular}{>{\raggedright\arraybackslash}p{3.1cm} >{\raggedright\arraybackslash}p{3.3cm} r}
\toprule
\textbf{Rule} & \textbf{Condition} & \textbf{Adj.} \\
\midrule
Protected Credit Risk ($Q_i$) &
  $N_i \geq N_{75}$ \textbf{and} $F_i = 1$ &
  $+0.08$ \\[4pt]
Recoverable Distress ($U_i$) &
  $L_i = 1$ \textbf{and} $T_i = 0$ &
  $+0.07$ \\[4pt]
Structural-Risk Penalty ($Z_i$) &
  $T_i = 1$ \textbf{and} $C_i < C_{25}$ &
  $-0.10$ \\[4pt]
Size-Burden Penalty ($K_i$) &
  $M_i > M_{75}$ \textbf{and} $C_i < C_{25}$ &
  $-0.06$ \\
\bottomrule
\end{tabular}
\end{table}

The thresholds $N_{75}$, $C_{25}$, and $M_{75}$ are the 75th, 25th,
and 75th percentiles of the corresponding normalized variable. During
cross-validation they are estimated on the training fold only
(Section~\ref{sec:leakage}), so the threshold that classifies a
held-out transaction never depends on that transaction's own value.
Rule $Q_i$ marks high credit risk that loss sharing transfers to the
FDIC; rule $U_i$ marks a deposit outflow without credit deterioration,
the pattern of a liquidity-driven failure; rules $Z_i$ and $K_i$
penalize a weakly capitalized buyer facing a deteriorating loan book
or a target that is large relative to itself. Government support is
not a condition of $U_i$: under the narrow definition it applies to
only one transaction in the panel and enters through $P_i$; a variant
that requires it is reported in Section~\ref{sec:results}.

The adjusted Spider Score is computed as:
\begin{equation}
S_i^* = S_i + 0.08\, Q_i + 0.07\, U_i - 0.10\, Z_i - 0.06\, K_i
\label{eq:adjusted}
\end{equation}
where each indicator $Q_i, U_i, Z_i, K_i \in \{0, 1\}$ according to
the conditions in Table~\ref{tab:interactions}.

\subsection{Final Rescaled Score}

Because $S_i^*$ may not fall in $[0, 1]$ across all observations,
it is rescaled as:
\begin{equation}
\tilde{S}_i = \frac{S_i^* - \min_{\text{train}}(S^*)}{\max_{\text{train}}(S^*) - \min_{\text{train}}(S^*)}
\label{eq:rescale}
\end{equation}
The bounds are estimated on the training split only and applied
unchanged to the test split, with test values clipped to $[0, 1]$, so
the test data never influence the normalization applied to themselves.
The rescaled score $\tilde{S}_i$ is the framework's decision index
(Models~1 and~4).

\section{Four-Model Comparative Framework}
\label{sec:models}

\subsection{Model 1: Spider Score with Fixed Thresholds (Baseline)}
\label{sec:model1}

Model~1 is the framework's model-free decision rule, designed for
interpretability and direct executive use. It applies fixed cut-offs
to the rescaled Spider score $\tilde S_i$ of
Eq.~\eqref{eq:rescale}:
\begin{itemize}
  \item \textbf{Pursue}: $\tilde S_i \geq 0.70$ --- low risk, strong value
  \item \textbf{Caution}: $0.40 \leq \tilde S_i < 0.70$ --- moderate risk
  \item \textbf{Avoid}: $\tilde S_i < 0.40$ --- high risk
\end{itemize}
Because $\tilde S_i$ involves no estimated coefficients, its ranking
ability can be assessed directly: Section~\ref{sec:results} reports the
AUC of $\tilde S_i$ itself as a score.

\subsection{Model 2: Logistic Regression (Statistical Benchmark)}
\label{sec:model2}

Model~2 is a logistic regression (LR) estimating the probability of a
positive announcement return from the eight inputs of Table~\ref{tab:variables}
(six continuous variables and the indicators $F_i$ and $H_i$). The
log-odds are
\begin{equation}
z = \beta_0 + \beta_1 X_1 + \beta_2 X_2 + \cdots + \beta_n X_n
\label{eq:logit_z}
\end{equation}
and the predicted probability of a positive announcement return is
\begin{equation}
\Pr(Y_i = 1 \mid \mathbf{X}_i) = \frac{1}{1 + e^{-z}},
\label{eq:logit_p}
\end{equation}
where $Y_i = 1$ denotes a positive announcement return
(Section~\ref{sec:outcome}). The inputs are standardized on the
training fold, and the coefficients carry a mild L2 penalty with
inverse strength fixed at $C = 1$, which stabilizes the estimates in a
sample of this size.

\subsection{Model 3: Spider Network Filter + Logistic Regression}
\label{sec:model3}

Model~3 extends Model~2 by adding the Spider Network features: the
rescaled score $\tilde S_i$ and the four interaction indicators
$Q_i, U_i, Z_i, K_i$, for 13 features in total. Its log-odds are
\begin{equation}
z_i = \beta_0 + \sum_{j} \beta_j X_{ij} + \gamma \tilde S_i +
      \sum_{k} \delta_k I_{ik},
\label{eq:model3}
\end{equation}
where $I_{ik}$, $k \in \{Q, U, Z, K\}$, are the four interaction
indicators of Table~\ref{tab:interactions}, and the predicted
probability follows Eq.~\eqref{eq:logit_p}. The L2 penalty strength is
chosen inside each training fold (Section~\ref{sec:leakage}).
The component scores ($V_i, P_i, R_i, B_i, E_i$) and the unrescaled
scores ($S_i$, $S_i^*$) are not added as separate predictors: they are
deterministic functions of the inputs, so including them would add
collinearity without new information.

Model~3 is compared against six further baselines under the identical
cross-validation protocol of Section~\ref{sec:leakage}
(Table~\ref{tab:main_results}): logistic regression with the least
absolute shrinkage and selection operator (LASSO, an L1 penalty); ridge
logistic regression; a decision tree; a random forest; extreme gradient
boosting (XGBoost); and a logistic regression with all 28 pairwise
interactions of the eight inputs. Ridge logistic regression is not a
different model family from Model~2: both are L2-penalized logistic
regressions on the same eight inputs, but Model~2 fixes the penalty at
$C = 1$, whereas the ridge comparator chooses it by inner
cross-validation, as Model~3 does. The ridge comparator therefore
isolates the contribution of the Spider features at a comparable level
of regularization. The
pairwise-interaction model is the most direct test of the framework:
it lets the data choose among all two-way interactions, whereas
Model~3 adds only the four interactions specified in
Table~\ref{tab:interactions}.

\subsection{Model 4: Cost-Sensitive Decision Policy}
\label{sec:model4}

Model~4 applies prescriptive Pursue/Caution/Avoid decision thresholds
to a transaction-level score. We evaluate three configurations rather
than treating a single set of cut-offs as given by assumption: (4A)
the framework's fixed thresholds applied to the rescaled Spider score
$\tilde S_i$, i.e.\ the rule of Model~1,
\begin{equation}
\text{Decision}_i =
\begin{cases}
\text{\bfseries Pursue}  & \text{score}_i \geq 0.70 \\
\text{\bfseries Caution} & 0.40 \leq \text{score}_i < 0.70 \\
\text{\bfseries Avoid}   & \text{score}_i < 0.40;
\end{cases}
\label{eq:model4}
\end{equation}
(4B) a configuration that maximizes the F1 score (the harmonic mean
of precision and recall), in which the Pursue threshold is
chosen from $\{0.55, 0.60, \ldots, 0.80\}$ inside each training fold,
using the inner grouped cross-validation predictions of Model~3, so
that the held-out transactions never influence it; and (4C) a
cost-sensitive configuration derived from a conservative 3:1
loss-to-benefit scenario. For 4B and 4C, ``score$_i$'' is Model~3's
out-of-fold predicted probability $\hat P(Y_i{=}1)$; 4A retains
$\tilde S_i$, so the comparison covers three deployable policies
rather than three thresholds applied to one score. For binary
evaluation, the conservative mapping is
\begin{equation}
\hat{Y}_i =
\begin{cases}
1 & \text{if Decision}_i = \text{\bfseries Pursue} \\
0 & \text{if Decision}_i \in \{\text{\bfseries Caution},\ \text{\bfseries Avoid}\}.
\end{cases}
\label{eq:model4_binary}
\end{equation}

In distressed M\&A, a false positive (recommending an acquisition that
the market then receives negatively) is more costly than a false
negative. We formalize this with a cost matrix in which $B$ is the
benefit of pursuing a transaction with a positive announcement return
and $L$ the loss from pursuing one with a negative return, and a
utility-adjusted score,
\begin{equation}
EU_i = \hat{P}(Y_i{=}1)\cdot B - (1-\hat{P}(Y_i{=}1))\cdot L,
\end{equation}
recommending Pursue only if $\hat P(Y_i{=}1) > L/(B+L)$. Under the
$L{:}B = 3{:}1$ scenario this gives a Bayes-optimal Pursue threshold of
0.75 for configuration~4C. The 3:1 ratio is a conservative scenario
rather than an externally documented industry parameter, so
Table~\ref{tab:sensitivity} sweeps it from 1:1 to 5:1. Because each row
of the sweep assumes a different loss, utilities at each row's own
ratio are not comparable across rows; we report each row's utility at
its own ratio, its gain over the better of the two trivial policies at
that ratio, and its utility at the common 3:1 ratio of
Table~\ref{tab:model4}. Realized utility is $(B \cdot TP - L \cdot
FP)/n$, the average payoff per transaction of the Pursue
recommendations, where $TP$ and $FP$ count the transactions with
positive and with negative announcement returns among those
recommended for pursuit; it is computed
on the actual outcomes with $B = 1$ and averaged over the 50 repeated
partitions. The results are discussed in
Section~\ref{sec:model4_results}.

\begin{table}[!t]
\caption{Model~4 Decision Policies ($n=55$; Mean over 50 Repeats; Utility with $B=1$, $L=3$)}
\label{tab:model4}
\centering
\footnotesize
\setlength{\tabcolsep}{2.5pt}
\begin{tabular}{lcccccc}
\toprule
\textbf{Policy} & \textbf{Thresh.} & \textbf{Pursue} & \textbf{TP/FP} & \textbf{F1} & \textbf{FPR} & \textbf{Util.} \\
\midrule
4A: Fixed ($\tilde S$) & 0.70/0.40 & 3.1 & 2.9/0.2 & 0.151 & 0.011 & 0.041 \\
4B: F1-tuned & 0.55--0.80 & 40.2 & 27.6/12.6 & 0.734 & 0.629 & $-$0.184 \\
4C: Cost (3:1) & 0.75/0.45 & 11.8 & 7.9/3.9 & 0.332 & 0.196 & $-$0.071 \\
\midrule
\multicolumn{7}{l}{\textit{Trivial policies}} \\
Pursue all & --- & 55 & 35/20 & 0.778 & 1.000 & $-$0.455 \\
Pursue none & --- & 0 & 0/0 & 0.000 & 0.000 & 0.000 \\
\bottomrule
\end{tabular}
\par\vspace{2pt}\parbox{0.94\columnwidth}{\scriptsize Pursue: transactions recommended for pursuit; TP/FP: of these, transactions with a
positive and with a negative announcement return; FPR: share of the 20 negative-CAR transactions recommended; Util.: realized
utility $(TP - 3\,FP)/55$.}
\end{table}

\begin{table}[!t]
\caption{Utility Sensitivity to the Assumed Loss-to-Benefit Ratio (Model~3 Probabilities; Mean over 50 Repeats)}
\label{tab:sensitivity}
\centering
\footnotesize
\setlength{\tabcolsep}{2.2pt}
\begin{tabular}{lcccccccc}
\toprule
 & & & & & & \multicolumn{2}{c}{\textbf{Own ratio}} & \textbf{3:1} \\
\cmidrule(lr){7-8}\cmidrule(lr){9-9}
\textbf{Ratio} & \textbf{Thr.} & \textbf{Pursue} & \textbf{TP/FP} & \textbf{FPR} & \textbf{F1} & \textbf{Util.} & \textbf{Gain} & \textbf{Util.} \\
\midrule
1:1 & 0.50 & 45.6 & 30.8/14.7 & 0.737 & 0.765 & 0.292 & 0.020 & $-$0.244 \\
2:1 & 0.67 & 24.3 & 17.6/6.8 & 0.339 & 0.590 & 0.073 & 0.073 & $-$0.051 \\
3:1 & 0.75 & 11.8 & 7.9/3.9 & 0.196 & 0.332 & $-$0.071 & $-$0.071 & $-$0.071 \\
4:1 & 0.80 & 7.0 & 4.3/2.8 & 0.138 & 0.197 & $-$0.123 & $-$0.123 & $-$0.073 \\
5:1 & 0.83 & 5.1 & 2.7/2.4 & 0.119 & 0.133 & $-$0.167 & $-$0.167 & $-$0.080 \\
\bottomrule
\end{tabular}
\par\vspace{2pt}\parbox{0.94\columnwidth}{\scriptsize Threshold $L/(B+L)$ applied to Model~3's out-of-fold probabilities. Gain: utility
minus that of the better trivial policy (pursue all or pursue none) at the same ratio.}
\end{table}

\subsection{Model Comparison}

Table~\ref{tab:model_compare} summarizes the key properties of all
four models.

\begin{table}[!t]
\caption{Comparative Summary of Four Models}
\label{tab:model_compare}
\centering
\footnotesize
\setlength{\tabcolsep}{4pt}
\begin{tabular}{>{\raggedright\arraybackslash}p{1.1cm} >{\raggedright\arraybackslash}p{2.5cm} >{\raggedright\arraybackslash}p{1.8cm} >{\raggedright\arraybackslash}p{2.0cm}}
\toprule
\textbf{Model} & \textbf{Purpose} & \textbf{Output} &
\textbf{Interpretability} \\
\midrule
Model 1 & Decision baseline & Spider score & Very high \\
Model 2 & Statistical benchmark & Probability & Moderate \\
Model 3 & Hybrid predictive model & Probability & Moderate to high \\
Model 4 & Executive decision rule & Category & Very high \\
\bottomrule
\end{tabular}
\end{table}

\subsection{Evaluation Protocol and Leakage Control}
\label{sec:leakage}

With 55 observations, a single train/test split produces noisy metric
estimates, so every model is evaluated by repeated cross-validation:
50 repetitions of 5-fold cross-validation, each with a different random
partition. Within a repetition, the out-of-fold predictions of the five
folds are pooled before the AUC, the Brier
score, and the other metrics are computed; we report the mean over the
50 repetitions and the 5th to 95th percentile range across them. The
repetitions reuse the same 55 transactions, so this range describes
sensitivity to the partition, not sampling uncertainty.

Some failed banks were sold to the same acquirer on the same day, and
their transactions share one cumulative abnormal return (CAR) of the
acquirer (Section~\ref{sec:outcome}).
The largest such group is four subsidiaries of FBOP Corporation sold
to U.S. Bank on October~30, 2009; Downey Savings and PFF Bank \& Trust,
sold to U.S. Bank on November~21, 2008, form another. If such
transactions were split across training and test folds, the outcome of
a test transaction could be learned from an identical outcome in the
training fold. We therefore treat each pair of listed parent and
event day (day~0, Section~\ref{sec:outcome}) as a cluster and use
\emph{grouped} stratified folds, in
which all transactions of a cluster fall in the same fold. The 55
transactions form 51 clusters.

Within \emph{every} fold, the following quantities are recomputed
using only the training rows of that fold and then applied without
modification to the held-out rows: (i) the standardization of the
inputs and the min-max normalization bounds; (ii) the quartile
thresholds that define $L_i$ and $T_i$ and the rule thresholds
$N_{75}$, $C_{25}$, $M_{75}$; (iii) the winsorization cap of $M_i$ and
the score-rescaling bounds of Eq.~\eqref{eq:rescale}; and (iv) the
model coefficients and hyper-parameters. Hyper-parameters (the penalty
strength $C \in \{0.1, 1, 10\}$ of the LASSO, ridge,
pairwise-interaction, and Spider models; the depth of the tree-based
models) are chosen by an inner grouped 3-fold cross-validation on the
training rows that minimizes the Brier score. Model~2 keeps $C = 1$.
Computing any of these quantities once on the full sample would let
test-fold information influence the features used to classify the
test fold.

Sampling uncertainty is assessed with three procedures, each of which
resamples or permutes the 51 event clusters rather than individual
transactions. The \emph{conditional bootstrap} resamples clusters
4,000 times, using each transaction's out-of-fold prediction averaged
over the 50 repetitions; it is conditional on those fitted
predictions and does not reflect the variability that comes from
refitting and tuning the models. The \emph{refitting bootstrap} draws
500 resamples of clusters and reruns the entire nested
cross-validation, including the inner tuning, on each resample (copies
of a cluster share its label and fall in the same fold); its intervals
therefore include that variability. The \emph{permutation test}
permutes outcomes across clusters and reruns the complete nested
procedure, including the inner tuning, on each of 400
permutations, averaging the AUC over three partitions per
permutation and using the same partition seeds for the observed data.
Its one-sided $p$-values are reported with and without a Holm
adjustment for the three models tested.

\subsection{Reproducibility}

The pipeline takes the source-verified transaction file as input and
produces, in sequence, the measured variables and the source ledger
(Section~\ref{sec:data}); the fold-wise Spider features for each of the
50 repeated partitions; the model comparison, paired comparison,
permutation tests, and both cluster bootstraps
(Tables~\ref{tab:main_results}, \ref{tab:paired}, and~\ref{tab:perm});
the ablation
(Table~\ref{tab:ablation}); the robustness checks, including the
bank-sector benchmark and the influence of single events
(Tables~\ref{tab:robustness} and~\ref{tab:risk}); the calibration and
SHAP figures
(Figs.~\ref{fig:calibration} and~\ref{fig:shap}); and the decision
analysis (Tables~\ref{tab:model4} and~\ref{tab:sensitivity}). Every
number reported in this paper is generated by this code from the
inputs; the code, the inputs, the source ledger, the software versions,
and checksums of every output are released together as one versioned
archive (Data and Code Availability).

\section{Data Structure and Variables}
\label{sec:data}

\subsection{Sample Construction}

The panel is built from the FDIC Failed Bank List~\cite{fdic_failed_list}
and BankFind Suite call-report data~\cite{fdic_bankfind}. Of 502
purchase-and-assumption failures, 469 have complete call-report data
for the failed bank in each of the four quarters before failure and
for the acquiring bank. We keep the 150 largest of these by
failed-bank total assets in the last quarter before failure, which
sets the size cut at \$351.5 million. In 94 of the 150 transactions
the acquiring bank's parent was a U.S. company listed on a U.S.
exchange; 44 acquirers were privately held, nine were owned by a
foreign bank, and three parents traded only over the counter. Daily
prices around the announcement are available for 55 of the 94; the
price data collected for this study do not cover the other 39, most
of whose parents were later merged or delisted
(Section~\ref{sec:limitations}). The primary panel therefore consists
of 55 transactions announced between 2008 and 2024. The same rules
without the size cut give 90 transactions with prices, out of 149 with
a U.S.-listed parent, which we use as a broader robustness panel.

\subsection{Dataset Design}

Each observation is one FDIC-assisted acquisition. The dependent
variable is \texttt{positive\_car} $\in \{0, 1\}$, where 1 indicates
a positive abnormal return to the acquirer's shareholders around the
announcement (Section~\ref{sec:outcome}). The data come from three public sources:
the FDIC Failed Bank List, which records every failure since October
2000 with the acquiring institution and the type of
transaction~\cite{fdic_failed_list}; the quarterly call reports of the
failed and acquiring banks in the FDIC BankFind
Suite~\cite{fdic_bankfind}; and the FDIC press release that announced
each transaction. SVB is one of the 55 transactions;
Section~\ref{sec:svb} uses it only to illustrate how the framework
scores a single acquisition.

\subsection{Variable Definitions and Source Verification}
\label{sec:variables}

Table~\ref{tab:variables} defines the eight inputs: six financial
inputs measured from call reports that were public before the
announcement, and two deal terms, whose availability
Section~\ref{sec:infoset} discusses. The failed bank's figures come
from its last public call report: the
latest quarter whose 30-day filing deadline fell before the failure
date. For 25 of the 55 banks, which failed within 30 days after a
quarter-end, this is the quarter before the last quarter-end, and the
FDIC's own press releases cite that earlier quarter's figures for such
banks; for the seven releases we compared, the release figures match
the inputs. The deposit change and the NPL-ratio change are measured
over the two quarters ending at that report. The acquirer's capital
ratio and total assets come from the acquiring bank's call report for
the last quarter-end at least 90 days before the failure, a report
that was public by the announcement and that cannot include the
acquired bank.

The two deal terms were coded from the FDIC press release of every
transaction, read for all 90 transactions of the broader panel; a
source ledger records, for each, the release number, its wording on
timing, the loss-share statement, and any bridge bank, systemic-risk
exception, or FDIC financing of the buyer. Forty of the 55 primary
transactions include an FDIC loss-share agreement. A release that
announces a loss-share transaction says so explicitly; where the
release is silent we code $F_i = 0$, and the acquirer's own filings
confirm this for each of the eight such transactions we checked.
Under the narrow definition used in the main analysis, government
support ($H_i = 1$) applies to SVB alone, which was resolved through a
bridge bank after a systemic-risk exception. A broad definition that
also counts FDIC financing of the buyer adds First Republic, whose
acquirer received \$50 billion of five-year fixed-rate financing; it is
used as a robustness check.

\begin{table*}[!t]
\caption{Variable Definitions, Network Roles, and Distribution in the Primary Panel ($n=55$)}
\label{tab:variables}
\centering
\footnotesize
\begin{tabular}{llll}
\toprule
\textbf{Variable} & \textbf{Description (timing)} & \textbf{Network role} & \textbf{Median [IQR] or count} \\
\midrule
$N$ & NPL ratio of the failed bank, noncurrent loans / loans (\%) (last public call report) & Risk & 13.96 [9.00, 17.23] \\
$G$ & Deposits of the failed bank, billions of USD (same report) & Value & 1.00 [0.50, 2.63] \\
$L$ & Change in the failed bank's deposits over the two quarters ending at that report (\%) & Risk (via $L_i$) & $-4.54$ [$-11.71$, 1.06] \\
$T$ & Change in the failed bank's NPL ratio over the same two quarters (percentage points) & Risk (via $T_i$) & 2.80 [0.70, 6.56] \\
$F$ & FDIC loss-share agreement (term of the bid; public with the FDIC's announcement) & Protection & 40 of 55 \\
$H$ & Bridge bank or systemic-risk exception (SVB only; public before its sale) & Protection & 1 of 55 \\
$C$ & Equity / total assets of the acquiring bank (\%) (last quarter-end $\geq 90$ days before failure) & Buyer strength & 9.35 [8.60, 11.63] \\
$M$ & Total assets of the failed bank / total assets of the acquiring bank & Execution & 0.13 [0.03, 0.27] \\
\midrule
$Y$ & Positive announcement return: $CAR_{[-1,+1]} > 0$, with day~0 the first session that could react & Target & 35 of 55 \\
\bottomrule
\multicolumn{4}{l}{\scriptsize IQR: interquartile range (25th and 75th percentiles).}\\
\end{tabular}
\end{table*}

\subsection{Information Set and Decision Time}
\label{sec:infoset}

The framework is designed for the acquiring bank. Its decision point is
the bid that the bank submits in the FDIC's sale process, before the
FDIC announces the winning bid, and its inputs fall into two groups
with different availability at that point. The six financial inputs
($N$, $G$, $L$, $T$, $C$, and $M$) are published information: they come
from call reports whose filing deadlines had passed before the failure,
so every bidder and every market participant could observe them before
the announcement. The two deal terms ($F$ and $H$) are not. In the
FDIC's sale process a failing bank is offered under alternative
transaction structures, with or without loss sharing, and each bid
specifies its structure~\cite{granja2017}; the winning bidder therefore
knows the loss-share terms of its own bid, and any FDIC financing it
negotiated, when it bids. Outside investors learn these terms only from
the FDIC's announcement, the release that starts the event window. The
narrow support indicator is an exception: it equals one only for SVB,
whose systemic-risk exception and bridge bank were announced on
March~12 and~13, 2023~\cite{jointstatement2023}, two weeks before its
sale. The broad indicator adds First Republic's FDIC financing, which
was disclosed with the transaction. Because bid documents are
confidential, we code $F$ and $H$ from the announcement and take the
announced terms to be those of the winning bid.

The models are therefore models of the bidder's decision problem: their
information set consists of published financial data and the bidder's
own transaction terms. They are not models of what outside investors
could predict before the announcement, for whom the deal terms are
part of the news. To show that the conclusions do not depend on
information that only the bidder had, Section~\ref{sec:results} also
reports every model restricted to the six published inputs, with $F$
and $H$ removed and the Spider score recomputed without its Protection
component and rule~$Q$.

\subsection{Outcome Variable Definition}
\label{sec:outcome}

The outcome variable is defined as
\begin{equation}
\texttt{positive\_car}_i = \mathbf{1}\bigl[CAR_{i,[-1,+1]} > 0\bigr],
\end{equation}
where $CAR_{i,[-1,+1]}$ is the cumulative abnormal return of the
acquirer over three trading days centered on the announcement:
\begin{equation}
AR_{i,t} = R_{i,t} - R_{m,t}, \qquad
CAR_{i,[-1,+1]} = \sum_{t=-1}^{+1} AR_{i,t}.
\label{eq:car}
\end{equation}
Here $R_{i,t}$ is the daily return of the acquiring bank's listed
parent, computed from closing prices adjusted for splits and
dividends, and $R_{m,t}$ is the return of the SPDR S\&P~500
exchange-traded fund (SPY). A positive CAR means that the market's
initial assessment of the acquisition was favorable. It is not a
measure of the acquisition's realized success, which would require
post-acquisition performance data, so we describe transactions with
$Y_i = 1$ as positively received at announcement rather than as
value-creating.

Day~0 is the first trading session in which the market could act on
the FDIC's announcement of the transaction. The timing of every
announcement was checked in the FDIC press release. Fifty-three of the
55 were released after the market close: 52 on a Friday evening and
Washington Mutual on a Thursday evening. Each of these releases places
the closing in the evening of the release date (typically stating that
the bank was closed that day and that the FDIC hotline is open ``this
evening''), and the three that carry a distribution time were sent
between 4:31 and 6:17~p.m.\ Eastern time.
For these transactions day~0 is the next trading session, usually a
Monday. First Republic was closed and sold before the market opened on
Monday, May~1, 2023, so day~0 is that Monday. SVB was closed on
March~10, 2023, and transferred to a bridge bank on March~13; its sale
to First Citizens was announced on Sunday, March~26~\cite{fdic2023svb},
so day~0 is Monday, March~27. This places the same three sessions around the news for
every transaction: the last session before it, the first session after
it, and the next one. It follows the event-study principle that day~0
is the first day on which prices can reflect the
news~\cite{mackinlay1997, berkman2009}; for a Friday-evening release it
coincides with the window $[0,+2]$ around the announcement date used by
Molyneux and Zhou~\cite{molyneux2022}. Cowan and
Salotti~\cite{cowan2015}, who note that the FDIC usually announces
closures on Friday evenings, report the windows $[0,+1]$ and $[-1,+1]$
around the announcement date, which we use as robustness checks.
Because many of the CARs are small, their sign can depend on the
window: 35 of the 55 transactions have a positive CAR under the
primary window, 39 under $[0,+1]$ around the announcement date, and 41 under
$[-1,+1]$ around it, which changes 14 labels relative to the primary
window. The two alternatives also cover the possibility that an
announcement reached the market before the close, since both contain
the announcement day itself.

The abnormal return is market-adjusted rather than estimated from a
market model. Over a three-day window the two approaches give very
similar results~\cite{brown1985, mackinlay1997}, and a market model
would have to be estimated in the months before each failure, which
for most of the panel coincide with the 2008--2010 banking crisis. As a
check, a market model estimated over trading days $-130$ to $-11$
labels 31 of the 55 transactions positive. Because the S\&P~500 does
not remove movements specific to bank stocks, which were large in
2008--2010, a second check replaces SPY with a bank-sector benchmark.
Our price data contain no bank index, so the benchmark is the
equal-weighted daily return of the 27 U.S.-listed bank holding companies
with complete price histories in the data, leaving out the acquirer and
any bank with its own announcement within four days; it labels 39 of
the 55 transactions positive. Acquirers owned by a
foreign bank (for example, TD Bank, Scotiabank de Puerto Rico, and
Compass Bank, then owned by BBVA) are excluded even when the parent's
shares or American depositary receipts trade in the United States: the
acquisition is small
relative to the parent, whose share price moves mainly with its home
market, so a CAR measured against the S\&P~500 says little about the
deal. Parents traded only over the counter are excluded because their
prices are thin. When one acquirer bought several failed banks on the
same day, those transactions share a single CAR;
Section~\ref{sec:leakage} accounts for this in the cross-validation.
Alternative outcomes, such as the 12-month change in return on assets,
deposit retention, and the FDIC's net resolution cost, are left for
future work.

\section{Case Study: SVB--First Citizens Acquisition}
\label{sec:svb}

SVB is one of the 55 panel transactions, and this section uses it to
illustrate how the Spider Network Filtering Framework scores a single
acquisition. It is not evidence of model performance, which
Section~\ref{sec:results} evaluates on the full panel.

\subsection{Background}

Silicon Valley Bank was founded in 1983 and grew to become the
primary banking institution for the U.S. venture capital and startup
ecosystem. At year-end 2022 it had total assets of about \$209
billion~\cite{fedreview2023}, which placed it among the 20 largest
U.S. banks. Its deposit base was highly concentrated among venture-capital-backed
technology and life-sciences firms---a characteristic that proved
decisive in its failure.

\subsubsection{Failure Mechanism}

SVB's failure was precipitated by a sequence of connected events
rather than a single cause. During 2020--2021, SVB deployed a surge
of deposits into long-duration held-to-maturity securities,
primarily agency mortgage-backed securities and U.S. Treasury bonds.
As the Federal Reserve began raising interest rates aggressively in
2022--2023, the market value of these securities fell sharply. At
year-end 2022 the held-to-maturity portfolio had an amortized cost of
\$91.3 billion and a fair value of \$76.2 billion, an unrealized loss
of about \$15 billion that accounting rules did not require SVB to
recognize~\cite{svb10k2022}. On March~8, 2023, SVB's parent announced
that it had sold \$21 billion of available-for-sale securities at an
after-tax loss of \$1.8 billion and would raise \$2.25 billion in
capital~\cite{fedreview2023}. A run began almost immediately, driven
by the interconnectedness of its depositor network: venture capital
firms communicated through shared channels and collectively
instructed portfolio companies to withdraw funds. SVB lost more than
\$40 billion in deposits on March 9, 2023, and expected to lose more
than \$100 billion the next day~\cite{fedreview2023}.

\subsubsection{FDIC Intervention and Acquisition}

The California Department of Financial Protection and Innovation
closed SVB on March 10, 2023, and appointed the FDIC as receiver. On
March~12, the Treasury, the Federal Reserve, and the FDIC announced
that all depositors would be protected under a systemic-risk
exception~\cite{jointstatement2023}, and the FDIC then transferred all
deposits and substantially all assets to the newly created Silicon
Valley Bridge Bank, N.A. On March 26, 2023,
the FDIC announced that First-Citizens Bank \& Trust Company, a
subsidiary of First Citizens BancShares, would assume all deposits
and loans of the bridge bank~\cite{fdic2023svb}. The transaction
included the purchase of approximately \$72 billion in assets at a
discount of \$16.5 billion and a loss-sharing agreement with the FDIC
covering commercial loans; approximately \$90 billion in securities
and other assets remained in FDIC receivership. First Citizens
assumed approximately \$56 billion in
deposits~\cite{firstcitizens2023}.

\subsection{Spider Network Scoring: SVB Transaction}

Table~\ref{tab:svb_scores} reports SVB's inputs and Spider scores. For
this illustration the continuous inputs are normalized with the bounds
of the full 55-transaction panel, and the rule thresholds are the
full-panel quantiles ($N_{75}$, $C_{25}$, and $M_{75}$ correspond to an
NPL ratio of 17.2\%, a buyer capital ratio of 8.60\%, and a relative
size of 0.27); in the cross-validation of Section~\ref{sec:results} the
same quantities are estimated on each training fold instead.

\begin{table}[!t]
\caption{Spider Network Inputs and Scores: SVB--First Citizens}
\label{tab:svb_scores}
\centering
\footnotesize
\begin{tabular}{lcc}
\toprule
\textbf{Component} & \textbf{Raw value} & \textbf{Model value} \\
\midrule
Deposits (Dec.~31, 2022) & \$175.4B & $G = 0.931$ \\
\textbf{Value score} $V$ & --- & \textbf{0.931} \\[3pt]
FDIC loss share & Yes & $F = 1$ \\
Government support & Bridge bank & $H = 1$ \\
\textbf{Protection score} $P$ & --- & \textbf{1.000} \\[3pt]
NPL ratio & 0.19\% & $N = 0.002$ \\
Deposit change, two quarters & $-7.8\%$ & $L = 0$ \\
NPL change, two quarters & $+0.06$ pp & $T = 0$ \\
\textbf{Risk score} $R$ & --- & \textbf{0.001} \\[3pt]
Buyer capital (Sep.~30, 2022) & 9.24\% & $C = 0.318$ \\
\textbf{Buyer strength} $B$ & --- & \textbf{0.318} \\[3pt]
Relative size & 1.92 & $M = 1.000$ \\
\textbf{Execution score} $E$ & --- & \textbf{1.000} \\
\bottomrule
\end{tabular}
\end{table}

Applying Eq.~\eqref{eq:raw},
\begin{align}
S_{\text{SVB}} &= 0.30\,V + 0.20\,P + 0.20\,B - 0.20\,R - 0.10\,E \notag\\
               &= 0.443.
\label{eq:svb_raw}
\end{align}
No interaction rule is triggered. The Protected Credit Risk rule
requires a high NPL ratio, and SVB's loan book was clean. The
Recoverable Distress rule requires a deposit outflow ($L = 1$): the run
that brought SVB down began on March~9, 2023, after its last public
call report, in which deposits had fallen 7.8\% over two quarters, short
of the panel's bottom quartile ($-11.7\%$). The public data available
before the failure did not show the run. The Structural-Risk Penalty
cannot apply because $T = 0$, and the Size-Burden Penalty, although SVB
was almost twice the size of First-Citizens Bank, requires a weakly
capitalized buyer, which First-Citizens was not. The adjusted and
rescaled scores are
\begin{equation}
S^*_{\text{SVB}} = 0.443, \qquad \tilde S_{\text{SVB}} = 1.000,
\label{eq:svb_adj}
\end{equation}
the highest of the panel, which places the transaction in the Pursue
zone of Model~1: its deposit franchise and full protection outweigh
the execution burden of a target larger than the buyer.

The market's assessment agreed. First Citizens' shares rose 53.7\% on
March~27, 2023, the first trading day after the acquisition was
announced~\cite{cnbc2023}; the three-day CAR around that day is 54.3\%,
and the bank later recorded a preliminary after-tax gain on the
acquisition of about \$9.8 billion~\cite{firstcitizens2023}. A single
transaction cannot validate or refute the framework, however, and
Section~\ref{sec:results} shows that the Spider score does not rank
the panel as a whole in the same order as the market.

\section{Results}
\label{sec:results}

\subsection{Main Model Comparison}
\label{sec:main_results}

Table~\ref{tab:main_results} reports mean performance over the 50
repeats of grouped 5-fold cross-validation for the eight models of
Section~\ref{sec:models}, together with the Spider score used directly
as a ranking (Model~1). Within each repeat the out-of-fold predictions
of the five folds are pooled before the metrics are computed; the
bracketed ranges show how much a result depends on the partition.

\begin{table*}[!t]
\caption{Model Comparison: 50 Repeats of Grouped 5-Fold Cross-Validation ($n=55$, 51 Event Clusters)}
\label{tab:main_results}
\centering
\footnotesize
\begin{tabular*}{\textwidth}{@{\extracolsep{\fill}}lcccc}
\toprule
\textbf{Model} & \textbf{AUC [5th, 95th pct.]} & \textbf{Event-weighted AUC} & \textbf{Brier [5th, 95th pct.]} & \textbf{Brier skill} \\
\midrule
L2 logistic, $C=1$ (Model~2) & 0.615 [0.535, 0.672] & 0.551 & 0.238 [0.218, 0.266] & $-$2.7\% \\
LASSO logistic & 0.518 [0.416, 0.620] & 0.479 & 0.249 [0.226, 0.282] & $-$7.6\% \\
L2 logistic, $C$ tuned (ridge) & 0.632 [0.556, 0.684] & 0.575 & 0.225 [0.211, 0.249] & 2.9\% \\
Decision tree & 0.585 [0.448, 0.675] & 0.533 & 0.291 [0.240, 0.362] & $-$25.9\% \\
Random forest & 0.605 [0.545, 0.666] & 0.569 & 0.231 [0.216, 0.244] & 0.1\% \\
XGBoost & 0.604 [0.534, 0.656] & 0.573 & 0.243 [0.223, 0.267] & $-$5.0\% \\
LR + pairwise interactions & 0.632 [0.560, 0.691] & 0.568 & 0.216 [0.200, 0.233] & 6.8\% \\
Spider + LR (Model~3) & 0.594 [0.525, 0.662] & 0.529 & 0.232 [0.214, 0.258] & $-$0.3\% \\
\midrule
Spider score alone (Model~1) & 0.423 [0.400, 0.457] & 0.450 & --- & --- \\
\bottomrule
\end{tabular*}
\par\vspace{2pt}\parbox{0.96\textwidth}{\scriptsize Means over the 50 repeats of the pooled out-of-fold AUC and Brier score;
brackets give the 5th and 95th percentiles across repeats (partition sensitivity, not a confidence interval).
Model~2 and the ridge comparator are the same L2-penalized logistic regression on the eight inputs; Model~2 fixes
$C=1$, the ridge comparator chooses $C$ by inner cross-validation. Event-weighted AUC collapses transactions that
share a CAR into one observation. Brier skill is the improvement over predicting the training-fold share of
positive-CAR transactions (Brier 0.231). The Spider score is not a probability, so only its ranking is evaluated.}
\end{table*}

Ridge regression and the logistic model with pairwise interactions
rank highest, with a mean AUC of 0.632 each, followed by
Model~2 (0.615), the random forest (0.605), XGBoost
(0.604), and Model~3 (0.594); the decision tree
(0.585) and the LASSO model (0.518) do worse. The ranges
across partitions overlap widely. Counting each event once lowers every
AUC, to between 0.479 and 0.575. Much of the
transaction-level discrimination comes from a single event: the four
FBOP subsidiaries sold to U.S. Bank on one day share a negative outcome
and are ranked near the bottom of the panel by Model~2, ridge
regression, the pairwise model, and Model~3 (mean percentile rank
below 0.08), while the remaining transactions with positive and
negative CARs receive similar predictions; without this cluster,
Model~2's mean AUC falls from 0.615 to 0.498
(Section~\ref{sec:robustness}). The pairwise-interaction model has the
lowest Brier score (0.216) and ridge regression the next
lowest (0.225); the Brier scores of Model~3 (0.232)
and Model~2 (0.238) are no better than that of predicting the
training-fold share of positive outcomes (0.231).
The Spider score itself has a mean AUC of 0.423: used as a
ranking, it places transactions with a positive CAR below the others
somewhat more often than above them.

\subsection{Paired Comparison and Chance-Level Tests}

\begin{table*}[!t]
\caption{Paired Comparison of Model~3 with Selected Baselines (Same Partitions, 50 Repeats)}
\label{tab:paired}
\centering
\footnotesize
\begin{tabular*}{\textwidth}{@{\extracolsep{\fill}}lccccc}
\toprule
 & \multicolumn{3}{c}{\textbf{50 partitions}} & \multicolumn{2}{c}{\textbf{Cluster bootstrap, AUC diff. [95\% interval]}} \\
\cmidrule(lr){2-4}\cmidrule(lr){5-6}
\textbf{Comparison} & \textbf{AUC diff. [5th, 95th pct.]} & \textbf{M3 higher} & \textbf{Brier diff. [5th, 95th pct.]} & \textbf{Conditional} & \textbf{Refitting} \\
\midrule
M3 $-$ Model~2 ($C=1$) & $-$0.020 [$-$0.085, $+$0.016] & 28\% & $-$0.006 [$-$0.024, $+$0.017] & $-$0.044 [$-$0.104, $+$0.001] & [$-$0.155, $+$0.131] \\
M3 $-$ Ridge ($C$ tuned) & $-$0.038 [$-$0.092, $+$0.010] & 10\% & $+$0.008 [$-$0.009, $+$0.028] & $-$0.056 [$-$0.115, $-$0.010] & [$-$0.166, $+$0.117] \\
M3 $-$ LR + pairwise & $-$0.038 [$-$0.085, $+$0.032] & 22\% & $+$0.017 [$-$0.004, $+$0.037] & $-$0.070 [$-$0.214, $+$0.049] & --- \\
M3 $-$ Random forest & $-$0.011 [$-$0.079, $+$0.076] & 44\% & $+$0.001 [$-$0.019, $+$0.027] & $-$0.050 [$-$0.188, $+$0.066] & --- \\
M3 $-$ XGBoost & $-$0.010 [$-$0.083, $+$0.055] & 42\% & $-$0.011 [$-$0.031, $+$0.010] & --- & --- \\
\bottomrule
\end{tabular*}
\par\vspace{2pt}\parbox{0.96\textwidth}{\scriptsize Differences are Model~3 minus the comparator on the same partitions; a negative Brier
difference favors Model~3. M3 higher: share of the 50 repeats in which Model~3's AUC is higher. Both bootstraps resample the 51 event
clusters. The conditional bootstrap (4,000 draws) uses each transaction's out-of-fold prediction averaged over the 50 repeats and
is conditional on those fitted predictions; the refitting bootstrap (500 draws) reruns the whole nested
cross-validation, including tuning, on every resample.}
\end{table*}

Table~\ref{tab:paired} compares Model~3 with selected baselines on
identical partitions. Model~3's AUC is lower than Model~2's by
0.020 on average and higher in only 28\% of the repeats; it is
lower than ridge regression's by 0.038 (higher in 10\%)
and than the pairwise model's by 0.038 (higher in 22\%),
and close to the random forest's and XGBoost's. Its Brier score is
slightly lower than Model~2's and XGBoost's but higher than ridge
regression's and the pairwise model's. These comparisons hold the 55
transactions fixed. The conditional cluster bootstrap, which resamples
the 51 event clusters but keeps each transaction's repeat-averaged
out-of-fold prediction, gives intervals of roughly $\pm 0.2$ around
each model's AUC; in it, the paired difference between Model~3 and
ridge regression, $-$0.056 (95\% CI [$-$0.115, $-$0.010]), excludes zero,
and the difference with Model~2, $-$0.044 ([$-$0.104, $+$0.001]), borders it.
Because these intervals hold the fitted predictions fixed, they omit
the variability of refitting and tuning the models. The refitting
bootstrap, which reruns the nested cross-validation on each of 500
resamples of clusters, gives wider intervals: from about 0.25 to
0.82 for each model's AUC, $-$0.022 ([$-$0.166, $+$0.117]) for
the difference with ridge regression, and $-$0.014 ([$-$0.155, $+$0.131]) for
the difference with Model~2. Both difference intervals contain zero,
although the
difference is negative in 62\% and 58\% of
the resamples. The Spider features do not improve ranking, but the
data cannot establish that they make it worse.

\begin{table}[!t]
\caption{Permutation Tests: Is Discrimination Above Chance?}
\label{tab:perm}
\centering
\footnotesize
\setlength{\tabcolsep}{2.5pt}
\begin{tabular}{lcccccc}
\toprule
 & & \multicolumn{2}{c}{\textbf{Null AUC}} & \multicolumn{3}{c}{\textbf{$p$-value}} \\
\cmidrule(lr){3-4}\cmidrule(lr){5-7}
\textbf{Model} & \textbf{AUC} & \textbf{Mean} & \textbf{95th} & \textbf{1-sided} & \textbf{Holm} & \textbf{2-sided} \\
\midrule
Model~2 ($C=1$) & 0.635 & 0.479 & 0.659 & 0.08 & 0.16 & 0.24 \\
Ridge ($C$ tuned) & 0.668 & 0.478 & 0.653 & 0.04 & 0.13 & 0.14 \\
Model~3 ($C$ tuned) & 0.600 & 0.477 & 0.647 & 0.12 & 0.16 & 0.36 \\
Spider score (Model~1) & 0.421 & 0.499 & 0.630 & 0.82 & --- & 0.36 \\
\bottomrule
\end{tabular}
\par\vspace{2pt}\parbox{0.94\columnwidth}{\scriptsize Outcomes permuted across the 51 event clusters (400 permutations). On every
permutation the complete nested procedure, including the inner tuning of $C$, is rerun; AUC is the mean over three
partitions, with the same partition seeds for the observed data. One-sided: AUC above the null; Holm: one-sided
$p$ adjusted for the three models tested; two-sided: distance from 0.5.}
\end{table}

Table~\ref{tab:perm} asks whether any model ranks transactions better
than chance. Outcomes are permuted across the 51 event clusters, which
preserves the shared outcomes within a cluster, and the complete nested
procedure, including the inner tuning of ridge regression and Model~3,
is rerun on each of 400 permutations, with the AUC averaged over
three partitions drawn with the same seeds as for the
observed outcomes. Under permutation the mean AUC of each logistic
model is about 0.48 rather than one half, a small-sample bias
of cross-validation, so the observed AUCs are compared with the
permutation distribution rather than with 0.5. On these partitions the
observed AUCs of Model~2 (0.635), ridge regression
(0.668), and Model~3 (0.600) are reached or exceeded
by 8\%, 4\%, and 12\% of the permuted
runs (one-sided $p = 0.08$, 0.04, and 0.12). After a Holm
adjustment for the three models, the smallest $p$-value is 0.13.
With 51 events, the evidence that any model ranks transactions better
than chance is therefore weak, and there is none that the Spider
features add to it. The Spider score's AUC below one half is also
within the range of chance (two-sided $p = 0.36$).

\subsection{Ablation Study}

\begin{table}[!t]
\caption{Ablation Study: 50 Repeats of Grouped 5-Fold Cross-Validation}
\label{tab:ablation}
\centering
\footnotesize
\setlength{\tabcolsep}{4pt}
\begin{tabular}{lccc}
\toprule
\textbf{Stage} & \textbf{AUC} & \textbf{Event AUC} & \textbf{Brier} \\
\midrule
A: measured inputs & 0.632 & 0.575 & 0.225 \\
B: A $+$ Spider score & 0.630 & 0.573 & 0.225 \\
C: A $+$ interaction rules & 0.598 & 0.534 & 0.231 \\
D: full SNFF (Model~3) & 0.594 & 0.529 & 0.232 \\
E: D without protection inputs & 0.615 & 0.556 & 0.228 \\
F: D without buyer capital & 0.632 & 0.573 & 0.225 \\
\bottomrule
\end{tabular}
\par\vspace{2pt}\parbox{0.94\columnwidth}{\scriptsize All stages are L2 logistic regressions with the penalty chosen in the inner
cross-validation; stage A is the ridge model of Table~\ref{tab:main_results}. In E and F the Spider score is recomputed
without the removed inputs and the rules that use them.}
\end{table}

Table~\ref{tab:ablation} isolates the contribution of each part of the
framework. Adding the Spider score to the measured inputs (stage~B) leaves the
mean AUC essentially unchanged (0.630 against 0.632 for stage~A),
whereas adding the four interaction rules (stage~C) lowers it to 0.598,
below stage~A in 92\% of the partitions; the full specification
(stage~D, Model~3) inherits this loss (0.594). Removing the protection
inputs and rule $Q$ from the full specification (stage~E) recovers part
of the loss (0.615), and removing buyer capital together with rules $Z$
and $K$ (stage~F) recovers all of it (0.632), which suggests that the
two rules conditioned on weak buyer capital add the most noise. No
stage improves on the measured inputs alone.

\subsection{Robustness}
\label{sec:robustness}

\begin{table*}[!t]
\caption{Robustness: Mean AUC over 50 Repeats under Alternative Outcomes, Specifications, and Sample}
\label{tab:robustness}
\centering
\footnotesize
\begin{tabular*}{\textwidth}{@{\extracolsep{\fill}}lccccc}
\toprule
\textbf{Variant} & \textbf{Positive CAR} & \textbf{Model~2} & \textbf{Ridge} & \textbf{Model~3} & \textbf{Spider score} \\
\midrule
Primary specification & 35/55 & 0.615 & 0.632 & 0.594 & 0.423 \\
Outcome: $CAR_{[0,+1]}$ around announcement date & 39/55 & 0.441 & 0.508 & 0.507 & 0.473 \\
Outcome: $CAR_{[-1,+1]}$ around announcement date & 41/55 & 0.572 & 0.601 & 0.591 & 0.425 \\
Outcome: market-model CAR & 31/55 & 0.642 & 0.648 & 0.647 & 0.409 \\
Outcome: CAR against bank-sector index & 39/55 & 0.277 & 0.280 & 0.413 & 0.505 \\
Published inputs only (no $F$, $H$) & 35/55 & 0.635 & 0.646 & 0.615 & 0.466 \\
Broad support definition ($H$) & 35/55 & 0.612 & 0.632 & 0.596 & 0.421 \\
Rule $U$ requires support (broad $H$) & 35/55 & 0.612 & 0.632 & 0.601 & 0.396 \\
Rule $Z$ removed & 35/55 & 0.615 & 0.632 & 0.593 & 0.429 \\
\midrule
Broader panel (no size cut) & 56/90 & 0.616 & 0.632 & 0.601 & 0.459 \\
\bottomrule
\end{tabular*}
\par\vspace{2pt}\parbox{0.96\textwidth}{\scriptsize Variants with the primary outcome use the partitions of Table~\ref{tab:main_results};
each alternative outcome uses 50 grouped partitions with the same seeds, stratified on its own labels. Published inputs only:
the six call-report inputs, without the deal terms $F$ and $H$; the Spider score is recomputed without its Protection component and
rule~$Q$. The broader panel of 90 transactions (78 event clusters) uses its own 50 grouped partitions.}
\end{table*}

Table~\ref{tab:robustness} repeats the comparison under alternative
outcomes, inputs, definitions, and samples. The outcome definition
matters more than any modeling choice. With the window $[0,+1]$ around
the announcement date, which labels 39 transactions positive, every
model's mean AUC is close to one half (0.441 for Model~2, 0.508 for
ridge regression, and 0.507 for Model~3). With $[-1,+1]$ around the
announcement date (41 positive), the AUCs are slightly lower than in
the primary specification (0.572, 0.601, and 0.591), and under the
market-model outcome (31 positive) the three models perform alike
(0.642 to 0.648). Under these three alternatives Model~3 ranks above
Model~2, by 0.005 to 0.065, and level with or slightly below ridge
regression, by 0.001 to 0.010. Against the bank-sector benchmark
(39 positive), every model ranks the transactions worse than
chance (0.277 for Model~2, 0.280 for ridge regression,
and 0.413 for Model~3). Ten labels change under this benchmark,
among them those of the four FBOP transactions, which turn positive
while Model~2 still ranks them near the bottom of the panel (mean
percentile rank 0.10). The same
cluster carries most of the discrimination in the primary
specification: without it, Model~2's mean AUC is 0.498 under
the primary outcome and 0.418 under the bank-sector
benchmark, against 0.615 and 0.277 with it. An AUC
below one half in cross-validation means that the relations fitted on
the training folds reverse in the held-out folds, as happens when the
fitted relations are mostly noise; the permutation analysis shows the
same downward bias. Relative to ridge regression, the Spider features
raise the Brier score in the primary specification (by 0.008) and lower
it under the four alternative outcomes (by 0.001 to 0.006), small
changes of inconsistent sign.

Restricted to the six published inputs, without the deal terms $F$ and
$H$ (Section~\ref{sec:infoset}), the models rank about as well as with
them, or slightly better (0.635 for Model~2, 0.646 for
ridge regression, and 0.615 for Model~3), so none of the
conclusions depends on information that only the bidder had. The broad
support definition, the variant of rule~$U$ that requires government
support, and the removal of rule~$Z$ change Model~3's mean AUC by less
than 0.01. In the broader panel of 90 transactions, Model~3 (0.601)
again ranks below Model~2 (0.616) and ridge regression (0.632), while
the pairwise-interaction model (0.676) and the random forest (0.660)
rank highest; because this panel contains the primary panel, it is not
an independent test. The Spider score used directly has a mean AUC
below one half in every variant (0.396 to 0.473) except the
bank-sector benchmark (0.505), and in the broader panel
(0.459).

\Figure[!t](topskip=0pt, botskip=0pt, midskip=0pt)[width=0.95\columnwidth]{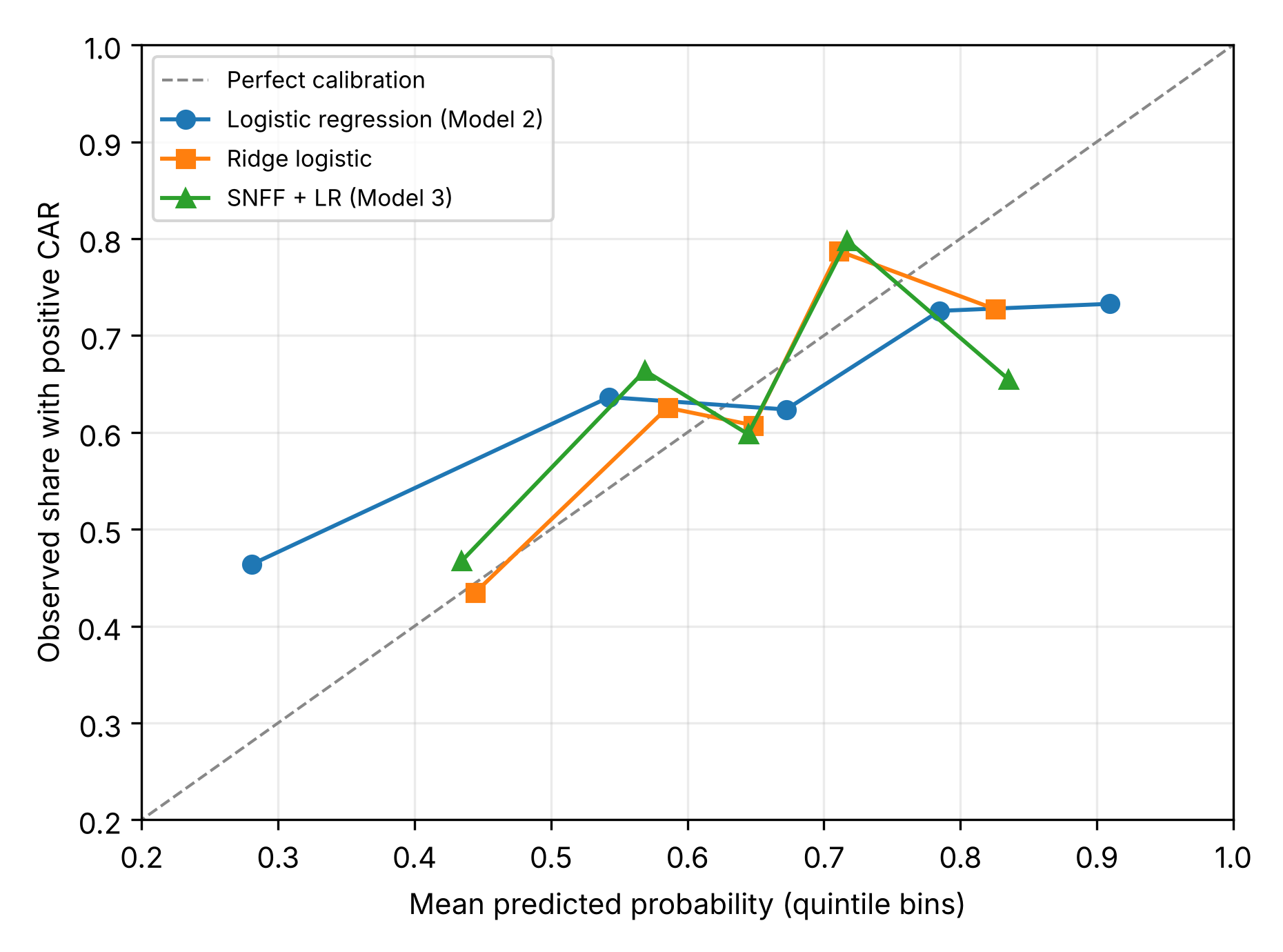}
{Calibration (reliability) of out-of-fold predicted probabilities,
pooled over the 50 repeats and grouped into five quantile bins, for
Model~2, ridge regression, and Model~3 ($n=55$). The dashed diagonal
indicates perfect calibration.\label{fig:calibration}}

\subsection{Calibration}

Fig.~\ref{fig:calibration} shows the calibration of the out-of-fold
probabilities, pooled over the 50 repeats. All three models match the
overall share of transactions with a positive CAR (mean predictions of
about 0.64 against an observed share of 0.64), but their predictions
are more dispersed than the outcomes warrant. For Model~2, the lowest
and highest quintiles of predicted probability average 0.280 and
0.910, while the observed shares in them are 0.464 and 0.733. Ridge
regression and Model~3 shrink the predictions toward the base rate,
with quintile means from about 0.44 to 0.83, but the observed share
does not rise steadily across their quintiles: Model~3's highest
quintile, with a mean prediction of 0.835, contains a lower share of
positive outcomes (0.655) than its fourth quintile (0.798).
The models separate the transactions more sharply than the outcomes
support, which is consistent with their weak discrimination and with
Brier scores close to that of the base rate.

\subsection{Explainability and the Direction of the Spider Components}

\Figure[!t](topskip=0pt, botskip=0pt, midskip=0pt)[width=0.95\columnwidth]{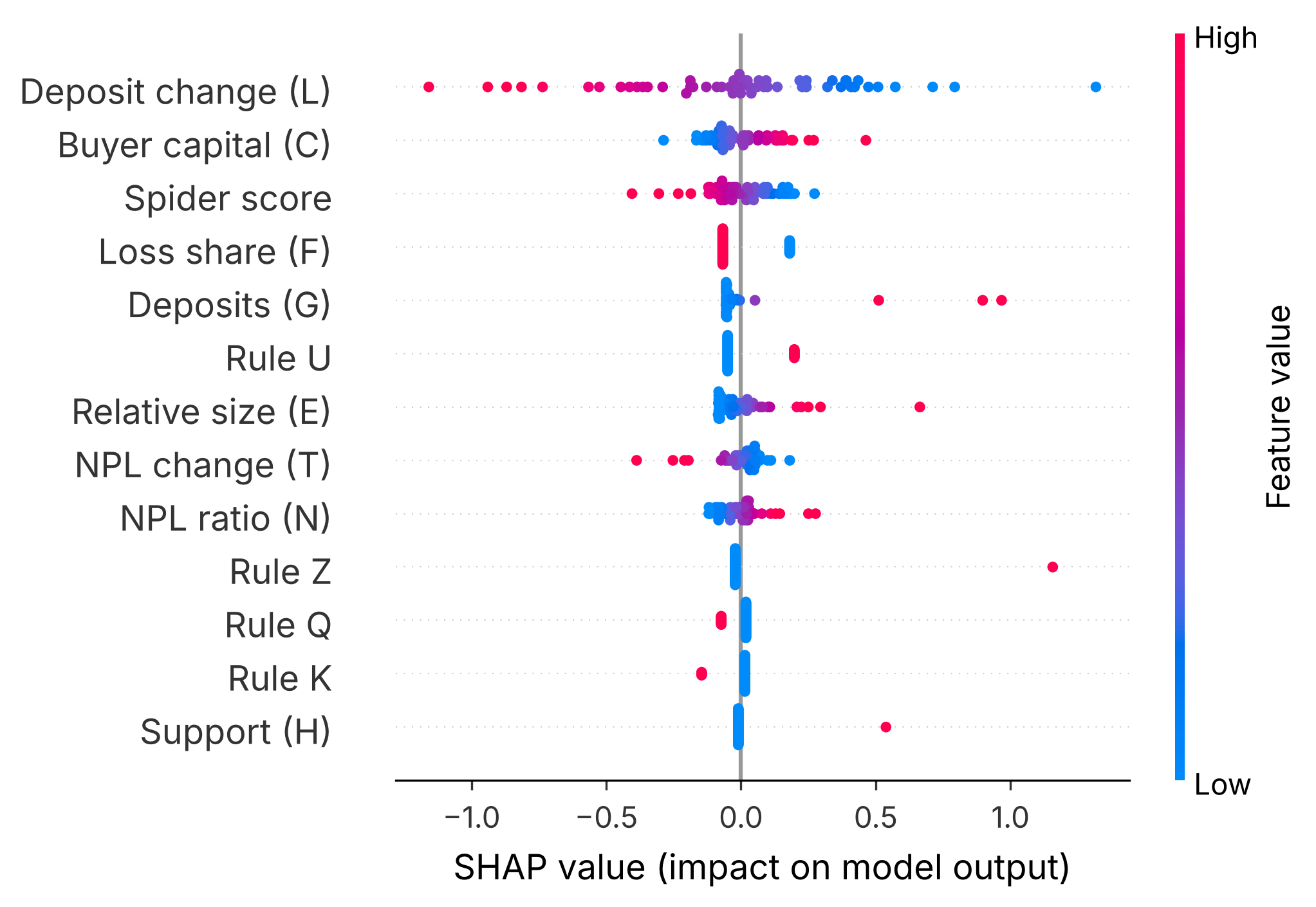}
{SHAP summary for Model~3 fitted on all 55 transactions (descriptive,
not out-of-sample). Each point is one transaction's contribution to
the log-odds of a positive CAR; color indicates the feature's value
(red = high, blue = low).\label{fig:shap}}

\begin{table}[!t]
\caption{Direction of the Spider Score and Its Components (AUC against a Positive CAR)}
\label{tab:components}
\centering
\footnotesize
\setlength{\tabcolsep}{3pt}
\begin{tabular}{lccccc}
\toprule
 & & \multicolumn{2}{c}{\textbf{Primary panel (55)}} & \multicolumn{2}{c}{\textbf{Other 35 deals}} \\
\cmidrule(lr){3-4}\cmidrule(lr){5-6}
\textbf{Component} & \textbf{Sign in $S$} & \textbf{AUC} & \textbf{$p$} & \textbf{AUC} & \textbf{$p$} \\
\midrule
Spider score $\tilde S$ & $+$ & 0.414 & 0.35 & 0.507 & 0.96 \\
Value $V$ & $+$ & 0.533 & 0.72 & 0.464 & 0.31 \\
Protection $P$ & $+$ & 0.454 & 0.54 & 0.464 & 0.71 \\
Buyer strength $B$ & $+$ & 0.537 & 0.66 & 0.714 & 0.06 \\
Risk $R$ & $-$ & 0.699 & 0.04 & 0.724 & 0.04 \\
Execution $E$ & $-$ & 0.600 & 0.29 & 0.675 & 0.30 \\
\bottomrule
\end{tabular}
\par\vspace{2pt}\parbox{0.94\columnwidth}{\scriptsize Normalization bounds and thresholds fitted on the primary panel only; no outcome
is used. ``Other 35 deals'' are the broader-panel transactions outside the primary panel. AUC above 0.5 means
higher component values go with a positive announcement return (SPY-adjusted, primary window); $p$ is two-sided, from
10,000 permutations of outcomes across event clusters. Table~\ref{tab:risk} repeats the Risk row under other benchmarks.}
\end{table}

A SHAP analysis of Model~3 fitted on the full sample
(Fig.~\ref{fig:shap}) is descriptive only, since the cross-validation
shows that the fitted relationships do not generalize reliably. It
ranks the failed bank's deposit change as by far the most influential
input, with a negative coefficient: larger deposit outflows before
failure go with a higher predicted probability of a positive CAR.
Buyer capital, the Spider score, the loss-share indicator, and the
deposit base follow, each with between a quarter and a third of that
weight. The Spider score and the loss-share indicator both carry
negative coefficients. Of the interaction rules, only $U$ carries a
weight comparable to that of the inputs; $Q$ and $K$ contribute little,
and rule $Z$, which fires for a single transaction in the full sample,
cannot be interpreted. The fitted signs largely agree with
Table~\ref{tab:components}: the model rewards deposit outflows and high
NPL ratios, two of the distress signals that the framework penalizes,
and gives the Spider score a negative weight.

Table~\ref{tab:components} examines the score directly. Because the
Spider score and its components involve no estimated coefficients,
their AUC against the outcome needs no cross-validation; the
normalization bounds and thresholds are fitted on the primary panel and
applied unchanged to the 35 transactions of the broader panel that are
outside it. Value, Protection, and Buyer Strength enter the score with
a positive sign but are only weakly related to a positive CAR in the
primary panel. Risk and Execution difficulty enter with a negative
sign, yet both have AUCs above one half against the primary outcome.
For the Risk component the association is the strongest of the six
(AUC 0.699, $p = 0.038$) and recurs in the other 35
transactions (AUC 0.724, $p = 0.042$). In those 35
transactions Buyer Strength is also associated with a positive CAR
(AUC 0.714, $p = 0.058$), in the direction the framework
assumes, but not in the primary panel. Of the transactions in the
lowest third of the Risk score, 7 of 19 have a positive CAR (median
CAR $-0.9\%$), against 14 of 18 in each of the other two thirds
(median CAR $+2.6\%$ and $+2.5\%$).

\subsection{Sensitivity of the Risk Component's Direction}
\label{sec:riskcheck}

\begin{table}[!t]
\caption{Sensitivity of the Risk Component's Association with Announcement Returns}
\label{tab:risk}
\centering
\footnotesize
\setlength{\tabcolsep}{3pt}
\begin{tabular}{lcc}
\toprule
\multicolumn{3}{l}{\textit{A. Benchmark and window: AUC (two-sided $p$)}} \\
\textbf{Outcome} & \textbf{Primary 55} & \textbf{Other 35} \\
\midrule
SPY, reaction day (primary) & 0.699 (0.04) & 0.724 (0.04) \\
Bank-sector index, reaction day & 0.514 (0.90) & 0.561 (0.66) \\
Market model, reaction day & 0.644 (0.14) & 0.637 (0.22) \\
SPY, $[0,+1]$ around release date & 0.423 (0.47) & 0.572 (0.57) \\
SPY, $[-1,+1]$ around release date & 0.343 (0.13) & 0.562 (0.62) \\
\midrule
\multicolumn{3}{l}{\textit{B. Influence in the primary panel: AUC}} \\
\textbf{Sample} & \textbf{SPY} & \textbf{Bank index} \\
\midrule
All 55 transactions & 0.699 & 0.514 \\
Leave one event out (range) & 0.639--0.731 & 0.494--0.573 \\
Without SVB, First Republic, WaMu & 0.736 & 0.540 \\
Without the five largest $|CAR|$ events & 0.733 & 0.518 \\
Announced 2008--2010 ($n=37$) & 0.717 & 0.442 \\
Announced 2011--2024 ($n=18$) & 0.679 & 0.711 \\
\bottomrule
\end{tabular}
\par\vspace{2pt}\parbox{0.94\columnwidth}{\scriptsize AUC of the Risk score $R$ against a positive CAR; $p$ from 10,000 permutations of
outcomes across event clusters. Normalization bounds and thresholds are fitted on the primary panel. The bank-sector index is the
equal-weighted return of 27 U.S.-listed bank holding companies, excluding the acquirer and banks with their own announcement
within four days. Panel B drops, in turn, each of the 51 event clusters, the three largest transactions, and the five event
clusters with the largest absolute CAR under each benchmark, and splits the panel by announcement year.}
\end{table}

The association of the Risk component with the primary outcome was
found after the six components had been examined, so
Table~\ref{tab:risk} tests it against the other outcome definitions and
against influential transactions. It depends on the benchmark. Against
the bank-sector benchmark of Section~\ref{sec:outcome}, the Risk
component's AUC is 0.514 ($p = 0.90$) in the primary
panel and 0.561 ($p = 0.66$) in the other 35
transactions; under the market model it is 0.644
($p = 0.14$), and with the two windows around the announcement
date it falls below one half in the primary panel (0.423 and
0.343). Against SPY, the association does not rest on a single
event or on the largest transactions: leaving out one event cluster at
a time gives AUCs from 0.639 to 0.731, and
leaving out SVB, First Republic, and Washington Mutual, or the five
event clusters with the largest absolute CAR, gives 0.736 and
0.733. Against the bank-sector benchmark, the same exclusions
give AUCs between 0.494 and 0.573, and the
association differs by period: it is reversed among the 37
transactions announced in 2008--2010 (0.442) and positive
among the 18 announced later (0.711). Against that
benchmark, 12 of the 19 transactions in the lowest third
of the Risk score have a positive CAR, against 14 and
13 of 18 in the other two thirds. Over the whole panel,
the association with the primary outcome therefore reflects the
benchmark: it appears when the acquirer's return is measured against
the broad market, which includes the movements of bank stocks, and not
when it is measured against other banks. We do not treat it as a
finding.

\subsection{Model 4 Decision-Threshold Results}
\label{sec:model4_results}

Table~\ref{tab:model4} reports the realized utility of the decision
policies of Section~\ref{sec:model4}. Pursuing every transaction
attains the highest F1 score of any policy (0.778), simply because 35
of the 55 transactions have a positive CAR, yet it realizes the lowest
utility ($-0.455$); pursuing none realizes zero. Configuration~4B,
tuned for F1, recommends Pursue on 40.2 of 55 transactions on
average, approves 62.9\% of the acquisitions with a negative
CAR, and realizes a utility of $-$0.184. Configuration~4C, the
cost-based threshold, pursues 11.8 transactions with a
false-positive rate of 19.6\% and realizes $-$0.071: Model~3's
probabilities do not discriminate well enough for the Bayes threshold
to pay off. Configuration~4A realizes a small positive utility
(0.041) because it recommends only about three transactions per
repeat, almost always the three with the highest Spider scores in the
panel---SVB, First Republic, and Washington Mutual---all of which had
a positive CAR; at this sample size that is a statement about three
large transactions rather than evidence of a deployable rule. The
sweep in Table~\ref{tab:sensitivity} shows that under symmetric costs
the model's decisions barely improve on pursuing every transaction
(gain 0.020), that at 2:1 they improve on both trivial policies
(gain 0.073), and that from 3:1 upward they lose to declining
every transaction; evaluated at the common 3:1 loss, every threshold in
the sweep has negative utility.

\subsection{Choice of Metrics and Assessment of H1}

Because a false positive costs more than a false negative, the
predictive comparison emphasizes AUC-ROC, which measures how well a
model ranks transactions, and the Brier score, which rewards
well-calibrated probabilities; the decision layer is evaluated
separately by realized utility and false-positive rate. F1 carries
little information here: a constant rule that predicts a positive CAR
for every transaction attains F1 = 0.778 at this sample's base rate,
above the F1 of every model at the conventional 0.5 threshold.

H1 is not supported. Model~3 does not improve on Model~2 in ranking
(AUC 0.594 against 0.615, higher in 28\% of the
partitions) and improves on it in calibration only marginally (Brier
0.232 against 0.238), while ridge regression and the
pairwise-interaction model score better than it on both criteria.
Under the two announcement-date windows and the market model
(Table~\ref{tab:robustness}), Model~3 does rank above Model~2 and has a
lower Brier score, which taken alone would favor H1; but ridge
regression, which differs from Model~2 only in its tuned penalty, ranks
as high or higher and accounts for most of the reduction in the Brier
score, so the improvement comes mainly from regularization rather than
from the Spider features. Against the bank-sector benchmark no model
ranks better than chance. Without the deal terms and under the support
and rule variants, Model~3 remains below Model~2 and ridge regression,
and in the broader panel it again ranks below both.

\section{Discussion}
\label{sec:discussion}

With the financial inputs taken from call reports published before
each failure, every deal term and announcement time checked against the
FDIC's own releases, and a validation design that keeps transactions
with a shared outcome in the same fold, none of the eight models
predicts reliably which FDIC-assisted acquisitions the market will
receive positively. Mean AUCs range from 0.518 to 0.632
and fall to 0.479--0.575 when each event counts once;
most of Model~2's discrimination comes from a single event of four
transactions; permutation tests that repeat the full model selection
give at most weak evidence that the best of the models ranks better
than chance (Table~\ref{tab:perm}); and with the window $[0,+1]$
around the announcement date or the bank-sector benchmark, every
model's discrimination falls to or below chance level
(Table~\ref{tab:robustness}). The Spider features do not improve on
logistic models that use the same inputs: in the primary specification
Model~3 ranks below Model~2 and below ridge regression in most
partitions, although the refitting bootstrap cannot exclude that the
three rank equally well; under the alternative windows and the market
model it never ranks above ridge regression, which uses Model~2's
inputs with a tuned penalty. The conclusions do not depend on the deal
terms, which only the bidder knows before the announcement: without
them, the models rank about as well as with them.

Three features of the problem plausibly limit what any model can
achieve here. The sample contains 51
independent events. The outcome is the sign of a three-day return,
and many of the returns are close to zero, so 14 of the 55 labels
change with the choice of window. And much of the value of an assisted
acquisition is set in the FDIC auction itself, through the price bid
and the terms of loss sharing, information that our inputs do not
contain beyond a binary loss-share indicator.

\subsection{The Direction of the Risk Penalty}

The Spider score subtracts the Risk component, as an acquirer
screening an open-market target would. The failed-bank literature
gives a reason to question that sign in FDIC auctions: the target's
distress is priced, through deeper discounts and loss-share coverage,
and part of the resulting wealth transfer can accrue to the winning
bidder~\cite{james1987, cowan2015, granja2017}. Measured against the
S\&P~500, our panel seems to agree, since the Risk component is
positively associated with a positive announcement return both in the
primary panel and in the 35 transactions outside it. The association
does not survive a bank-sector benchmark, and in the primary panel its
sign changes with the event window (Table~\ref{tab:risk}). The data
therefore do not settle
the direction of the risk penalty, in either direction: they give no
support to penalizing distress, and no reliable support to rewarding
it. Whether distress is priced in the acquirer's favor can only be
judged with the terms of the winning bid, the purchase discount and
the loss-share coverage, which our inputs contain only as a binary
loss-share indicator.

\subsection{Decision Analysis}

The Model~4 results in Section~\ref{sec:model4_results} carry the
paper's methodological lesson. A risk committee or board that tuned a
decision rule for F1 (Configuration~4B) would, on this sample, approve
most acquisitions that the market received negatively and realize a
utility well below
that of declining every transaction, even though pursuing every
transaction attains the highest F1 of any policy. The metric used to
tune a decision rule must match the asymmetric cost of the decision.
The converse does not hold here: the cost-based threshold
(Configuration~4C) also loses to declining every transaction, because
probabilities with little discriminating power cannot support a
demanding threshold, and the small gain of the fixed-threshold rule
(4A) rests on the three largest transactions in the panel.

\subsection{Regulatory and Practitioner Applications}

Table~\ref{tab:practitioners} summarizes how different users could
apply an interaction-aware decision framework of this kind. The
present evidence does not support using the Spider score to screen
transactions; its value for these users lies in making the reasoning
behind a recommendation explicit and auditable, so that a committee
can see which interactions a recommendation relies on and test them
against outcomes, as this paper does.

\begin{table}[!t]
\caption{Potential Users of an Interaction-Aware Decision Framework}
\label{tab:practitioners}
\centering
\footnotesize
\begin{tabular}{>{\raggedright\arraybackslash}p{2.3cm}>{\raggedright\arraybackslash}p{5.2cm}}
\toprule
\textbf{User} & \textbf{Use of the framework} \\
\midrule
Acquiring bank & Structuring the review of a distressed target \\
FDIC / regulators & Comparing bidders and transaction structures \\
Risk committee & Auditing the drivers of a recommendation \\
Board / executives & Pursue, Caution, or Avoid deliberations \\
Investors & Assessing the sources of acquisition gains \\
\bottomrule
\end{tabular}
\end{table}

\subsection{Directions for Future Work}

Five extensions follow from the results. First, the purchase discount
and the loss-share coverage of the winning bid, which the FDIC reports
in its bid summaries for many transactions, would let the framework
account for the price the acquirer pays for the target's risk.
Second, the sample can be enlarged: 39 of the 94 transactions with a
U.S.-listed acquirer among the 150 largest failures lack price data
here, mostly because the acquirer was later merged or delisted, and a
price source that covers delisted securities would recover many of
them; a larger sample would also allow the association between target
distress and announcement returns to be tested with adequate power
under several benchmarks. Third, the expert-assigned weights could be
estimated, or updated as due-diligence information arrives, within an
explicit probabilistic model of the framework; doing so requires such
a specification and stage-level data that the current panel does not
contain. Fourth,
industry-derived variables, such as deposit velocity in the weeks
before failure, depositor concentration (the vulnerability illustrated
by SVB), key-person dependency in commercial lending, and core-system
compatibility, may carry information that quarterly call reports
miss, and resolutions under other regimes, such as the 2017 sale of
Banco Popular Espa\~{n}ol to Santander under the EU Single Resolution
Mechanism, would show whether the interaction rules carry over to
different protection arrangements. Fifth, outcomes that measure
realized success, such as post-acquisition profitability, deposit
retention, or the FDIC's resolution cost, would complement the
announcement return, which records only the market's initial
assessment.

\subsection{Limitations}
\label{sec:limitations}

Several limitations should be noted. First, the panel has 55
transactions and 51 independent events, which limits statistical
power; the permutation tests and the refitting bootstrap, both of
which repeat the model fitting and selection, show that the data
barely separate the models from one another or from chance. Second,
the panel covers only acquirers with a U.S.-listed parent and
available prices: 39 of the 94 such transactions among the 150 largest
failures are missing, mostly because the parent was later merged or
delisted, so the panel over-represents acquirers that remained
independent. Third, the outcome is the sign of a three-day abnormal
return. It records the market's initial reception of the acquisition,
not its realized success, and its sign is sensitive to the window and
to the benchmark when returns are small: 14 of the 55 labels change
with the window, and 10 change when a bank-sector index replaces the
S\&P~500. Fourth, the deal terms are coded from the FDIC's
announcement and taken to be the terms of the winning bid; they belong
to the bidder's information set at the time of its bid, but not to that
of outside investors before the announcement. The inputs also omit the
price paid: deal-level discounts are not available for most
transactions, the loss-share indicator does not distinguish coverage
levels, and the support indicator applies to a single transaction.
Fifth, announcement timing is established from the wording of the
releases and, for a few of them, from their distribution time; the
alternative windows cover an announcement that reached the market
before the close. Sixth, the component weights and rule adjustments
are expert-assigned rather than estimated or elicited through the
Analytic Hierarchy Process, and the interaction graph is a conceptual
description of the rules rather than a structure that the model
learns from. Finally, the analysis is predictive, not causal.

\section{Conclusion}
\label{sec:conclusion}

This paper introduced the Spider Network Filtering Framework (SNFF),
an interpretable, graph-inspired feature-engineering scheme that
encodes conditional dependencies among the financial dimensions of a
distressed bank acquisition as four interaction rules, and evaluated it
on 55 FDIC-assisted acquisitions by U.S.-listed acquirers in which
every input, deal term, and announcement time was checked against
public sources. The models describe the bidder's decision, with
published financial data and the bidder's own transaction terms as
inputs, and the outcome is the market's initial reception of the
acquisition, not its realized success. Under grouped, nested
cross-validation the Spider-enhanced logistic model reached a mean AUC
of 0.594, against 0.615 for logistic regression on the same
inputs and 0.632 for ridge regression. Permutation tests that
repeat the complete model selection give at most weak evidence that any
of the models ranks better than chance, the refitting bootstrap does
not separate the Spider-enhanced model from either logistic baseline,
and the hypothesis that the Spider features improve ranking and
calibration (H1) is not supported. The Spider score itself does not
rank better than chance (AUC 0.423). An apparent positive
association between the target's measured distress and the acquirer's
announcement return, which would reverse the score's risk penalty,
holds against the S\&P~500 but not against a bank-sector benchmark, so
the data do not settle the sign of that penalty.

The acquisition of Silicon Valley Bank by First Citizens BancShares
served as an illustrative case, not as evidence: the framework placed
it at the top of the panel, in agreement with the market, but the
panel as a whole does not support the score.

The decision analysis shows that a rule tuned for F1 approves most
acquisitions that the market received negatively and realizes negative
utility when pursuing such an acquisition costs three times the
benefit of pursuing one that the market received positively, while
pursuing every transaction maximizes F1; at that cost ratio no policy
evaluated here improves materially on declining every transaction. In
distressed acquisition decisions, the evaluation metric is itself a
modeling decision with economic consequences. Future work should add
the terms of the winning bid, enlarge the sample with acquirers whose
shares are no longer traded, measure realized acquisition success, and
test the revised framework on transactions not used to design it.

\section*{Data and Code Availability}

The datasets, source documentation, analysis code, and reproducibility materials supporting this study are publicly available in the GitHub repository:

\url{https://github.com/malihealiasgari20/FDIC-SNFF-Decision-Framework}

The repository contains the primary 55-transaction panel, the broader 90-transaction panel, the FDIC press-release source-verification ledger documenting transaction terms and announcement timing, the bank-sector benchmark returns, and the Python code used to generate the statistical analyses, tables, and figures presented in this paper.

The materials constitute the Version 4.0 research replication package, dated October 30, 2026. The repository also provides the software dependencies, computational environment information, reproduction instructions, and SHA-256 checksums for verifying the integrity of the included research files.

The transaction-level banking and acquisition data were obtained from publicly available Federal Deposit Insurance Corporation (FDIC) sources, including the FDIC Failed Bank List, the BankFind Suite, and official FDIC press releases. Cumulative abnormal returns (CARs) were calculated using daily closing stock prices adjusted for stock splits and dividends. Historical price data were retrieved in June 2026 from Yahoo Finance using the \texttt{yfinance} Python library, with Financial Modeling Prep serving as an alternative source when necessary.

The repository includes the derived returns, reconstructed analytical datasets, and computational scripts necessary to reproduce the reported statistical analyses. The original historical stock-price series are not redistributed. Consequently, reconstruction of certain upstream market-return calculations requires separate access to the underlying historical price data.
\appendix
\section{Source Documentation for the Primary Panel}
\label{sec:appendix}

Table~\ref{tab:appendix} lists the 55 transactions of the primary panel
with the deal terms and announcement timing coded from the FDIC press
releases and the resulting outcome. For each transaction the release
announcing the purchase and assumption was read in the FDIC
press-release archive (\url{https://archive.fdic.gov}) or in the FDIC's
e-mail bulletins, which record the time of distribution. Several
institutions acquired in one FDIC transaction share a release: the four
FBOP subsidiaries in the panel and five others sold to U.S. Bank
(PR-195-2009), and Downey Savings and PFF Bank \& Trust (PR-124-2008).
PR-30-2024 is the release of April~26, 2024 on Republic First Bank.
CARs are shown in percent with two decimals; each outcome is computed
from the unrounded CAR. The complete ledger for the 90 transactions of
the broader panel, with the wording of each release on timing, loss
sharing, bridge banks, and FDIC financing, is part of the data
release.

\begin{table*}[!t]
\caption{Primary Panel: Source-Verified Terms, Timing, and Outcome of the 55 Transactions}
\label{tab:appendix}
\centering
\scriptsize
\setlength{\tabcolsep}{3pt}
\renewcommand{\arraystretch}{0.95}
\begin{tabular*}{\textwidth}{@{\extracolsep{\fill}}llllccccc}
\toprule
\textbf{Failed bank} & \textbf{Parent} & \textbf{Announced} & \textbf{FDIC release} & \textbf{Timing} & \textbf{LS} & \textbf{H} & \textbf{CAR (\%)} & \textbf{Y} \\
\midrule
Washington Mutual Bank & JPM & 2008-09-25 & PR-85-2008 & A & 0 & 0 & 9.52 & 1 \\
First Republic Bank & JPM & 2023-05-01 & PR-34-2023 & P & 1 & 0 & 1.77 & 1 \\
Silicon Valley Bank & FCNCA & 2023-03-26 & PR-23-2023 & N & 1 & 1 & 54.31 & 1 \\
Colonial Bank & BBT & 2009-08-14 & PR-143-2009 & A & 1 & 0 & $-$3.85 & 0 \\
Downey Savings and Loan Assoc., F.A. & USB & 2008-11-21 & PR-124-2008 & A & 1 & 0 & 0.71 & 1 \\
Westernbank Puerto Rico & BPOP & 2010-04-30 & PR-97-2010 & A & 1 & 0 & 3.33 & 1 \\
United Commercial Bank & EWBC & 2009-11-06 & PR-201-2009 & A & 1 & 0 & 53.68 & 1 \\
California National Bank & USB & 2009-10-30 & PR-195-2009 & A & 1 & 0 & $-$0.91 & 0 \\
Doral Bank & BPOP & 2015-02-27 & PR-24-2015 & A & 0 & 0 & 7.32 & 1 \\
Republic Bank & FULT & 2024-04-26 & PR-30-2024 & A & 0 & 0 & 6.73 & 1 \\
Franklin Bank, SSB & PB & 2008-11-07 & PR-113-2008 & A & 0 & 0 & 5.04 & 1 \\
Park National Bank & USB & 2009-10-30 & PR-195-2009 & A & 1 & 0 & $-$0.91 & 0 \\
First NBC Bank & HWC & 2017-04-28 & PR-34-2017 & A & 0 & 0 & 2.17 & 1 \\
PFF Bank \& Trust & USB & 2008-11-21 & PR-124-2008 & A & 1 & 0 & 0.71 & 1 \\
San Diego National Bank & USB & 2009-10-30 & PR-195-2009 & A & 1 & 0 & $-$0.91 & 0 \\
Eurobank & OFG & 2010-04-30 & PR-95-2010 & A & 1 & 0 & $-$5.23 & 0 \\
First Community Bank & USB & 2011-01-28 & PR-18-2011 & A & 0 & 0 & 1.99 & 1 \\
First Regional Bank & FCNCA & 2010-01-29 & PR-26-2010 & A & 1 & 0 & $-$1.83 & 0 \\
Integra Bank, N.A. & ONB & 2011-07-29 & PR-128-2011 & A & 1 & 0 & 9.52 & 1 \\
Pacific National Bank & USB & 2009-10-30 & PR-195-2009 & A & 1 & 0 & $-$0.91 & 0 \\
United Western Bank & FCNCA & 2011-01-21 & PR-13-2011 & A & 1 & 0 & 1.34 & 1 \\
Vineyard Bank, N.A. & ZION & 2009-07-17 & PR-125-2009 & A & 1 & 0 & $-$16.29 & 0 \\
Peoples First Community Bank & HBHC & 2009-12-18 & PR-234-2009 & A & 1 & 0 & 4.98 & 1 \\
County Bank & WABC & 2009-02-06 & PR-19-2009 & A & 1 & 0 & 6.54 & 1 \\
Temecula Valley Bank & FCNCA & 2009-07-17 & PR-126-2009 & A & 1 & 0 & $-$5.99 & 0 \\
Horizon Bank & WAFD & 2010-01-08 & PR-4-2010 & A & 1 & 0 & $-$3.19 & 0 \\
Alliance Bank & ZION & 2009-02-06 & PR-18-2009 & A & 1 & 0 & $-$3.46 & 0 \\
Integrity Bank & RF & 2008-08-29 & PR-74-2008 & A & 0 & 0 & 24.46 & 1 \\
Columbia River Bank & COLB & 2010-01-22 & PR-18-2010 & A & 1 & 0 & 0.53 & 1 \\
Guaranty Bank & FCNCA & 2017-05-05 & PR-37-2017 & A & 0 & 0 & $-$1.63 & 0 \\
Crescent Bank and Trust Company & RNST & 2010-07-23 & PR-164-2010 & A & 1 & 0 & 8.25 & 1 \\
Venture Bank & FCNCA & 2009-09-11 & PR-170-2009 & A & 1 & 0 & 0.22 & 1 \\
First Chicago Bank \& Trust & WTFC & 2011-07-08 & PR-117-2011 & A & 1 & 0 & 6.27 & 1 \\
San Joaquin Bank & CVBF & 2009-10-16 & PR-185-2009 & A & 1 & 0 & 4.54 & 1 \\
Fidelity Bank & HBAN & 2012-03-30 & PR-38-2012 & A & 0 & 0 & 0.84 & 1 \\
Colorado Capital Bank & FCNCA & 2011-07-08 & PR-118-2011 & A & 1 & 0 & 0.38 & 1 \\
TeamBank, N.A. & GSBC & 2009-03-20 & PR-46-2009 & A & 1 & 0 & 19.65 & 1 \\
Darby Bank \& Trust Company & ABCB & 2010-11-12 & PR-249-2010 & A & 1 & 0 & 5.14 & 1 \\
Haven Trust Bank & BBT & 2008-12-12 & PR-134-2008 & A & 0 & 0 & 9.82 & 1 \\
Warren Bank & HBAN & 2009-10-02 & PR-179-2009 & A & 0 & 0 & $-$1.45 & 0 \\
Bank of Whitman & COLB & 2011-08-05 & PR-132-2011 & A & 0 & 0 & $-$3.35 & 0 \\
K Bank & MTB & 2010-11-05 & PR-242-2010 & A & 1 & 0 & $-$2.11 & 0 \\
The First National Bank of Olathe & EFSC & 2011-08-12 & PR-134-2011 & A & 1 & 0 & $-$0.09 & 0 \\
Sun American Bank & FCNCA & 2010-03-05 & PR-43-2010 & A & 1 & 0 & 9.13 & 1 \\
Washington First International Bank & EWBC & 2010-06-11 & PR-133-2010 & A & 1 & 0 & 4.59 & 1 \\
The Park Avenue Bank & VLY & 2010-03-12 & PR-51-2010 & A & 1 & 0 & 3.24 & 1 \\
Security Savings Bank, F.S.B. & SFNC & 2010-10-15 & PR-226-2010 & A & 1 & 0 & $-$0.14 & 0 \\
Vantus Bank & GSBC & 2009-09-04 & PR-163-2009 & A & 1 & 0 & 4.78 & 1 \\
First Bank of Idaho, FSB & USB & 2009-04-24 & PR-60-2009 & A & 0 & 0 & $-$3.90 & 0 \\
Inter Savings Bank, FSB (InterBank) & GSBC & 2012-04-27 & PR-48-2012 & A & 1 & 0 & $-$0.98 & 0 \\
Valley Bank & GSBC & 2014-06-20 & PR-47-2014 & A & 0 & 0 & 2.81 & 1 \\
Bradford Bank & MTB & 2009-08-28 & PR-155-2009 & A & 1 & 0 & 0.04 & 1 \\
Mountain National Bank & FHN & 2013-06-07 & PR-50-2013 & A & 0 & 0 & 0.14 & 1 \\
Wheatland Bank & WTFC & 2010-04-23 & PR-90-2010 & A & 1 & 0 & $-$5.60 & 0 \\
Sun Security Bank & GSBC & 2011-10-07 & PR-159-2011 & A & 1 & 0 & 0.22 & 1 \\
\bottomrule
\end{tabular*}
\par\vspace{2pt}\parbox{0.96\textwidth}{\scriptsize Ordered by failed-bank total assets. Parent: ticker of the acquiring bank's listed parent (Hancock
Holding appears as HBHC and, for 2017, under its later ticker HWC; BBT is BB\&T, now Truist). Timing: A = announced after the
close (day~0 is the next session), P = before the open (same day), N = on a non-trading day (next session).
LS = FDIC loss-share agreement; H = bridge bank or systemic-risk exception (narrow definition). CAR = market-adjusted
$CAR_{[-1,+1]}$ around day~0; Y = 1 if the unrounded CAR is positive. Transactions bought by one acquirer on the same day share a CAR.}
\end{table*}


\EOD
\end{document}